\documentclass[journal,twoside]{IEEEtran}
\usepackage{cite}

\ifCLASSINFOpdf
   \usepackage[pdftex]{graphicx}
\else
\fi
\usepackage{amsmath,amssymb}
 
\DeclareMathOperator*{\argmin}{arg\hspace{1pt}min} 
\usepackage{url}

\usepackage{longtable}
\usepackage[noend]{algorithmic}
\usepackage{algorithm}
\usepackage{subfigure}
\usepackage{array}
\usepackage{textcomp}
\usepackage{stfloats}
\usepackage{verbatim}
\usepackage{cite}
\usepackage{color, xcolor, colortbl}
\usepackage{bm}
\usepackage{multirow}
\usepackage{multicol}
\usepackage{latexsym}
\usepackage{arydshln}
\usepackage{afterpage}
\usepackage{booktabs}
\usepackage[hidelinks]{hyperref}
\usepackage{pdfpages}

\usepackage{xr}
\definecolor{mygreen}{rgb}{0.93,0.96,0.82} 
\definecolor{mypink}{rgb}{0.98,0.91,0.92}
\definecolor{myyellow}{rgb}{0.99,0.95,0.72}

\newcommand{\win}{\cellcolor{mygreen}\bfseries} 
\newcommand{\worst}{\cellcolor{mypink}\it}

\definecolor{green2}{rgb}{ .23, .68, .46}

\newcommand{\cth}{\multicolumn{1}{c}}

\newcommand{\REMARK}[1]{\item[] \vspace{1mm} /* #1 */}

\newcommand{\our}{EPOS}
\newcommand{\ours}{\our~}

\def\BibTeX{{\rm B\kern-.05em{\sc i\kern-.025em b}\kern-.08em T\kern-.1667em\lower.7ex\hbox{E}\kern-.125emX}}

\begin{document}
%
\title{An Evolutionary Algorithm Assisted by an Ensemble of Pareto-Optimal Surrogate Models}
%
%
%

\author{Kei~Nishihara,~\IEEEmembership{Member,~IEEE,}
        Yaochu~Jin,~\IEEEmembership{Fellow,~IEEE,}
        and~Masaya~Nakata,~\IEEEmembership{Member,~IEEE}


\thanks{\copyright~2026 IEEE. Personal use of this material is permitted. Permission from IEEE must be obtained for all other uses, in any current or future media, including reprinting/republishing this material for advertising or promotional purposes, creating new collective works, for resale or redistribution to servers or lists, or reuse of any copyrighted component of this work in other works.}

\thanks{Received 7 November 2025; revised 9 February 2026 and 30 April 2026; accepted 7 June 2026. This work was supported in part by the Japan Society for the Promotion of Science (JSPS) KAKENHI under Grant 20H04254, Grant 22KJ1409, Grant 25K03195, and Grant 26K21330; and in part by the International Collaboration Fund for Creative Research of National Natural Science Foundation of China (NSFC ICFCRT) under Grant W2441019. This article was recommended by Associate Editor Y. S. Ong. (\textit{Corresponding author: Masaya Nakata.})}

\thanks{Kei Nishihara is with the Organization for the Promotion of Education, Yokohama National University, Yokohama 2408501, Japan (e-mail: nishihara-kei-pv@ynu.ac.jp).}
\thanks{Yaochu Jin is with the School of Engineering, Westlake University, Hangzhou 310030, China (e-mail: jinyaochu@westlake.edu.cn).}
\thanks{Masaya Nakata is with the Faculty of Engineering, Yokohama National University, Yokohama 2408501, Japan (e-mail: nakata-masaya-tb@ynu.ac.jp).}

\thanks{This article has supplementary material provided by the authors and color versions of one or more figures available at https://doi.org/10.1109/TCYB.2026.3702665.}
\thanks{Digital Object Identifier 10.1109/TCYB.2026.3702665}

}

%
%

\markboth{IEEE TRANSACTIONS ON CYBERNETICS (A preprint of the ``Accepted article'' version)}
{NISHIHARA \MakeLowercase{\textit{et al.}}: EA ASSISTED BY AN EPOS MODELS}
%



\maketitle

\begin{abstract}
    An ensemble of surrogate models helps improve the prediction quality and robustness of surrogate models, and in turn, the search performance of surrogate-assisted evolutionary algorithms (SAEAs). Although different degrees of smoothness of the approximated fitness landscapes need to be carefully designed for an effective ensemble, little attention has been paid to the explicit tuning of the degree of smoothness derived by surrogate models. This study proposes an adaptive ensemble SAEA, which automatically constructs plausible ensemble models by optimizing their parameter settings. Unlike existing adaptive/ensemble SAEAs, which consider prediction accuracy alone, the proposed algorithm optimizes the structure of radial basis function networks by solving bi-objective minimization problems of approximation error and model complexity, resulting in robust ensemble models of accurate surrogate models with different degrees of smoothness of the approximated fitness landscapes. As a result, the over/under-fittings are reduced. Additionally, an infill criterion is designed so that surrogate models with different degrees of smoothness can contribute to the solution prescreening. The experimental results demonstrated the statistical superiority of our algorithm over state-of-the-art SAEAs on a single-objective benchmark and real-world problem sets under an expensive optimization scenario. The source code of the proposed algorithm is available at \href{https://github.com/haranychan/EPOS}{https://github.com/haranychan/EPOS}.
\end{abstract}
    

\vspace{-1mm}

%
\IEEEpeerreviewmaketitle

\section{Introduction} \label{sec:int}

\IEEEPARstart{L}{ike} vehicle aerodynamic optimization \cite{Jin2001-sk,Le2013-gk,Dong2021-cs} and neural architecture search \cite{Baldeon_Calisto2021-jn,Lu2019-om}, the real world contains a wide range of expensive optimization problems (EOPs), wherein the function evaluations (FEs) are computationally and/or financially expensive. For example, the aerodynamic optimization of the rear of a car model requires one hour to evaluate a design using numerical simulations \cite{Le2013-gk}. In EOPs, the number of FEs is usually restricted to a few hundreds due to the limited computational and/or financial resources.

Surrogate-assisted evolutionary algorithms (SAEAs) \cite{Khaldi2024-hx} are a representative methodology to solve EOPs under a restricted number of FEs. Hereinafter, SAEAs for single-objective optimization problems (SOPs) are considered in this work if not described differently. Typically, SAEAs use machine learning (ML) techniques to construct surrogate models of objective functions and prescreen candidate solutions by estimating their fitness values with surrogate models. Therefore, the performance of SAEAs strongly depends on the quality of the surrogate models. Many efforts have been made to improve their quality in SAEAs having a single surrogate model. 
For example, to increase the prediction accuracy of surrogate models, SADE-ATDSC \cite{Nishihara2022-nr} adaptively selects training data of surrogate models and some SAEAs use dimensionality reduction techniques \cite{Yang2024-op}. IKAEA \cite{Zhan2021-wu} applies the incremental learning to the Kriging model \cite{Lophaven2002-bf} to reduce the time complexity from cubic to quadric. Zhao \textit{et al.} proposed a supervised SAEA mechanism using a fitness evaluation accuracy to maintain prediction accuracy of surrogate models and solution diversity~\cite{Zhao2023-zm}. For multi-objective optimization problems (MOPs), DA-PSL \cite{Lu2024-sd} employs the Pareto set learning to obtain Pareto-optimal solutions accurately and efficiently. TEA \cite{Zhang2024-xh} and DISK \cite{Zhang2025-fs} utilize the Pareto dominance relationship derived from the uncertainty of Kriging to preserve elite solutions.

However, relying on a single surrogate model can make SAEAs less robust against model deterioration owing to inevitable over/under-fittings \cite{Wang2023-ss}. To mitigate this risk while improving the prediction accuracy, ensemble approaches have been developed \cite{Jin2011-fd}. The key idea is to take advantage of the outputs from surrogate models with different degrees of smoothness of the approximated fitness landscapes \cite{Lim2007-hc} and aggregate the outputs, such as using the weighted average \cite{Goel2007-ya}. For example, GS-SOMA \cite{Lim2010-zh} uses different model types for an ensemble to utilize both advantages of ragged and smooth shapes of the approximated fitness landscape. This utilizes the different fidelities of surrogate models depending on their options like ML model types and their settings \cite{Jin2011-fd}. While a ragged landscape can raise prediction accuracy, smoothed landscapes make the search easier than using the original exact fitness landscape, also known as ``bless of uncertainty \cite{Lim2010-zh}.'' HESNFO \cite{Yu2022-wl} employs the bootstrap sampling to construct multiple radial basis function networks (RBFNs) \cite{Park1991-et}, and XGBEA \cite{Mao2024-hh} for MOPs adopts XGBoost \cite{Chen2016-hf} as surrogate models. These approaches also essentially vary the degree of smoothness of the approximated fitness landscapes. Hence, this degree of smoothness is crucial and should be carefully designed in ensemble SAEAs. To this end, two directions of efforts have been done to design ensemble surrogate models.
\begin{itemize}
    \item \textit{Ensemble SAEAs}. Beyond using one model type and perturbed training data, the degree of smoothness of the approximated fitness landscapes is purposefully varied \cite{Jin2011-fd,Guo2019-at}. ESAMA \cite{Lim2007-hc} and GS-SOMA construct surrogate models with different model types such as Kriging, RBFN, and polynomial regression (PR) \cite{Myers1995-cu}, then aggregate their outputs. EAPSO \cite{Li2019-ff} uses the RBFN surrogate models with different radial basis functions (RBFs) to diversify the approximated fitness landscape. Jin \textit{et al.} \cite{Jin2004-nh} prepared several artificial neural networks (ANNs) \cite{Amari1967-tp} having different structures and hyperparameters for the ensemble to improve prediction accuracy.

    \item \textit{Adaptive Ensemble SAEAs}. This approach adds an adaptation mechanism to further enhance the robustness of ensemble models against different problems since the performance of SAEAs depends on their surrogate model options \cite{Tong2021-ay}. 
    For example, ASMEA \cite{Yu2020-oz} identifies the most accurate surrogate models from five model types, labeling them as elites. Then, it chooses the best model among these elites and their ensemble model. HESNFO constructs five surrogate models such that each RBF differs and forms an ensemble model of the best two.
\end{itemize}

Although these studies have approached the goal of constructing robust surrogate models to some extent by ensembling surrogate models with different degrees of smoothness, they are often subject to the following important limitations. First, since it is difficult to measure and control the degree of smoothness directly, existing work has no choice but using a limited number of manually predefined surrogate model options like different model/RBF types for ensembles. This indirect control can decrease the robustness of ensemble models as the degree of smoothness cannot be handled as intended. Second, limited choices of ensemble elements lack the freedom of tuning the degree of smoothness. For example, since constructing ANNs is time-consuming, only few ensemble elements can be used \cite{Jin2004-nh}, although they may not be suitable for given situations. This results in occasional failures to balance the degrees of smoothness among ensemble elements and improve the performance of SAEAs. Finally, even ensemble SAEAs install adaptive mechanisms, adaptive ensemble SAEAs only consider the prediction accuracy as the selection criterion of surrogate model options \cite{Li2022-bl}, although the balance between convergence and diversity or uncertainty is frequently discussed in evolutionary algorithms (EAs) \cite{Ong2006-dk} or infill criteria \cite{Xie2024-qk}, respectively. However, depending on the prediction accuracy alone lacks diversity of surrogate models and their robustness. A surrogate model with the minimum error often fails to improve the performance of SAEAs~\cite{Field2008-vg}.

Accordingly, this study proposes an EA assisted by an Ensemble of Pareto-optimal Surrogate models (\our). We newly focus on the property that the degree of smoothness of the approximated fitness landscapes can be controlled with the structural/computational complexity in ANNs \cite{Jin2004-nh}, including RBFNs, which we use for surrogate models due to computational efficiency \cite{Tong2021-ay}. \ours is fundamentally distinguished from the existing ensemble SAEAs with few predefined ensemble elements in that \ours automatically constructs ensemble models of RBFNs having various but well-controlled degrees of smoothness by optimizing the RBFN structure during a run. Generally, although an ANN becomes more expressive as it has a more complex structure, the generalization performance can deteriorate \cite{Murata1999-be}.
This forms an MOP of the approximation error and model complexity with a tradeoff between them \cite{Jin2008-gl}. Thus, within ANN-based surrogate models, solving this MOP helps the flexible and elaborate design of the degree of smoothness while maintaining good prediction accuracy and robustness, unlike the existing adaptive SAEAs with accuracy-based selection alone. In \our, Nondominated Sorting Genetic Algorithm II (NSGA-II) \cite{Deb2002-vn} minimizes the aforementioned MOP to obtain Pareto-optimal surrogate models. Then, \ours ensembles them and prescreens offspring generated by Differential Evolution (DE) \cite{Storn1997-ny}. For the infill criterion, we emulated the lower confidence bound (LCB) by considering the difference in outputs of ensemble elements as uncertainty. This even enhances the value of our ensemble approach because \ours can utilize more solid uncertainty calculated from elaborate ensemble elements.

The main contributions of this work are as follows.
\begin{itemize}
    \item To the best of our knowledge, this study is the first to consider the minimization of both approximation error and model complexity simultaneously in the automatic design of ensemble surrogate models for SAEAs. Some work on the evolutionary neural architecture search \cite{Baldeon_Calisto2021-jn, Lu2019-om} considers this MOP. However, motivations differ from each other; while they intend to install appropriate ANNs for devices of varying scales, our work improves the quality of ensemble models for SAEAs. This study highlights the importance of considering model complexity alongside approximation error in surrogate model adaptation.
    \item We demonstrate that obtaining Pareto-optimal surrogate models is an effective approach to constructing a robust ensemble model. Unlike existing ensemble SAEAs that rely on predefined ensemble elements, \ours offers a high degree of freedom in surrogate model design through multi-objective optimization. It selects appropriate models from a diverse pool. Moreover, while recent surrogate models such as ELDR \cite{Harada2023-wp} and XGBoost used in ELDR-SAHO \cite{Harada2023-wp} and XGBEA, respectively, are becoming increasingly complex and hard to configure, our approach can automatically prepare suitable ensemble models, even if we use a simple ML technique, i.e., RBFN.
    \item To achieve a low-cost implementation of model uncertainty in the LCB infill criterion for robust solution selection, we emulated it using the difference among outputs of the ensemble members. This completed the establishment of our automatic surrogate modeling method.
\end{itemize}

The remainder of this work is organized as follows. Section~\ref{sec:rtw} introduces related work. Section~\ref{sec:tch} describes the components of \our, i.e., DE, NSGA-II, and RBFN. Section~\ref{sec:our} explains the detailed mechanism of \our. In Section~\ref{sec:exp}, we compare the performance of \ours with state-of-the-art SAEAs on the CEC 2020 bound-constrained SOP benchmark suite \cite{Yue2019-vz}. Section~\ref{sec:dis} discusses the adequacy of the \ours design by comparing the performance of \ours and its variants, and applies \ours to real-world problems. Finally, Section~\ref{sec:con} summarizes our work and discusses future directions.

\section{Related Work} \label{sec:rtw}

This section introduces SAEAs related to this work. TABLE~\ref{tab:rtw} lists some single-objective adaptive and/or ensemble SAEAs in chronological order of their proposals. The table summarizes the variation seed, number of variation seeds, and the used surrogate model to prescreen solutions for FEs or the used options, where ``variation seed'' refers to the adaptation targets or the ensemble elements that realize various surrogate models. In other words, the number of variation seeds is the number of candidate surrogate models considered for the adaptation or ensemble.

\begin{table*}[t]
    \caption{Adaptive SAEAs, ensemble SAEAs, and adaptive ensemble SAEAs as related work.}
    \vspace{-2mm}
    \label{tab:rtw}
    \centering
    \scalebox{0.83}{
        \begin{tabular}{c|rlcl}\hline\hline
            \multicolumn{1}{c|}{Classification} & \cth{Algorithm}               & \cth{Variation Seed}           			            & \begin{tabular}{c} Number \\[-0.5mm] of Seeds \end{tabular} & \cth{Used Surrogate Model or Option} 								\\\hline
            \multirow{14}{*}{\hspace{-2mm}\begin{tabular}{c} Adaptive \\[-1mm] SAEAs \end{tabular}}  
                                        & SUMO        \cite{Gorissen2009-av}    & Model type                               	            & 7     				& A surrogate with the minimum MSE                              \\
                                        & RBFBS       \cite{Li2019-db}          & Spread parameter in RBFN        			            & 11   	 				& A surrogate with the minimum RMSE                             \\
                                        & aRBF-NFO    \cite{Yu2021-nb}          & RBF type in RBFN                                      & 5     				& A surrogate with the minimum RMSE           		            \\
                                        & SAHO        \cite{Pan2021-jj}         & EA type                                               & 2     				& An EA that does not stagnate	                                \\
                                        & DDEA-MESS   \cite{Yu2022-iw}          & Infill criterion                                      & 3                     & A surrogate selected by scheduled probability                 \\
                                        & SADE-ATDSC  \cite{Nishihara2022-nr}   & Training data sampling criterion                      & 4    					& A surrogate with the minimum RMSE                             \\
                                        & ESA         \cite{Zhen2023-ta}        & Infill criterion                                      & 4                     & A surrogate selected by Q-learning                            \\
                                        & SAF-TD       \cite{Lin2024-sp}        & EA type and its hyperparameters                       & 4     		        & An EA with hyperparameters that does not stagnate             \\
                                        & AutoSAEA    \cite{Xie2024-qk}         & Combination of model type and infill criterion	    & 8    			        & A combination selected as a multi-armed bandit problem        \\
                                        & AKAEA        \cite{Lu2025-yy}         & Solutions and search strategies for subpopulations    & 2		                & Suitable ones depending on search situations                  \\
                                        & AutoSAPSO    \cite{Dai2025-jv}        & Model type, EA type, and its hyperparameters          & 10		            & \hspace{-2mm}\begin{tabular}{l} A surrogate with the better R2 score \\[-0.5mm] \ and an EA with hyperparameters that does not stagnate \end{tabular} \\
                                        & HSADE-CS     \cite{Yu2025-wr}         & Hyperparameters of DE variants                        & 3		                & Hyperparameters adapted by the extended method of \cite{Tanabe2014-el} \\
                                        & CoE-SAEA     \cite{Xie2025-qo}        & Combination of model type and infill criterion        & 8		                & A combination selected by a large language model              \\
                                        & $(\theta_{l}, \theta_{u})$-PMTO \cite{Wei2025-wb}& Task variation in multi-task optimizations & \ -${}^*$             & Task variations improving the coverage of the task space      \\
            \hline \multirow{12}{*}{\hspace{-2mm}\begin{tabular}{c} Ensemble \\[-1mm] SAEAs \end{tabular}}  
                                        & Jin \textit{et al.} \cite{Jin2004-nh} & Structure and parameters of ANN                       & 3-5                   & Bagging of ANN surrogates with different architectures        \\
                                        & ESAMA       \cite{Lim2007-hc}         & Model type                                            & 2                     & Bagging of three surrogates with different model types        \\
                                        & GS-SOMA     \cite{Lim2010-zh}         & Model type                                            & 3                     & Bagging of three surrogates with different model types        \\
                                        & LES-CDE     \cite{Jin2015-vy}         & Training data in overlapped local regions             & 3                     & Bagging of three surrogates with different training dataset   \\
                                        & SPOT2       \cite{Bartz-Beielstein2017-ok} & Model type                                       & 3                     & Stacking of three surrogates with different model types       \\
                                        & CAL-SAPSO   \cite{Wang2017-ku}        & Model type                                            & 3                     & Bagging of three surrogates with different model types        \\
                                        & EAPSO       \cite{Li2019-ff}          & RBF type in RBFN                                      & 2                     & Bagging of two RBFN surrogates with different RBFs            \\
                                        & SAEA-UGC    \cite{Liu2021-en}         & Model type                                            & 3                     & Bagging of three surrogates with different model types        \\
                                        & DS-SAEA     \cite{Ma2025-lq}          & RBF type in RBFN                                      & 3				        & Bagging of three surrogates with different RBF types          \\
                                        & SADE-DSDS   \cite{Liu2025-mc}         & Infill criterion                                      & 2 		            & Nondominated solutions of two infill criteria                 \\
                                        & CESAEA      \cite{Guo2025-nq}         & Classification models of distributed clients          & \ \ -${}^{**}$        & Boosting of the client surrogates for the server              \\
                                        & CCoEA-ASAA \cite{Wei2026-zw}          & Training data sampled via bootstrap and distributed client models & \ \ -${}^{**}$        & Bagging of the client surrogates for the server   \\
            \hline \multirow{5}{*}{\hspace{-2mm}\begin{tabular}{c} Adaptive \\[-1mm] Ensemble \\[-1mm] SAEAs \end{tabular}}  
                                        & ASMEA       \cite{Yu2020-oz}          & Model type                                            & 5                     & \hspace{-2mm}\begin{tabular}{l} Select a surrogate with the minimum RMSE from those \\[-0.5mm] \ with small RMSEs and their bagging model \end{tabular} \\
                                        & HESNFO      \cite{Yu2022-wl}          & RBF type in RBFN                                      & 5    					& Bagging of two surrogates with top two RMSEs                  \\
                                        & DSP-SAEA    \cite{Liu2023-ml}         & Model type (three and their bagging model)            & 4    			        & A surrogate with the minimum RMSE                             \\
                                        & PS-SAEA     \cite{Wang2025-jb}        & Training data sampled via bootstrap                   & \ \ \ -${}^{***}$     & Bagging of randomly selected surrogates with RMSEs            \\
            \cline{2-5}                 & \ours (Ours)                          & Spread parameter and the number of hidden layer nodes of RBFN  & 110          & Pareto-optimal surrogates obtained by NSGA-II                 \\
            \hline
        \end{tabular}
    }
    \\\vspace{1mm} 
    \hfill \scriptsize ${}^*$ It depends on the task situation. $(\theta_{l}, \theta_{u})$-PMTO is for expensive multi-task optimizations. \ ${}^{**}$ It depends on the number of clients. They are for EOPs with distributed data.
    \\
    \hfill \scriptsize ${}^{***}$ PS-SAEA is for offline EOPs and constructs 2,000 surrogates only at the beginning of the optimization using the given initial data. 
    \\
    \hfill \scriptsize The other SAEAs are for online EOPs with the centralized data, and ``Number of Seeds'' indicates the number of surrogates constructed in a generation.
    \vspace{-2mm}
\end{table*}

\subsection{Adaptive SAEAs}

Adaptive SAEAs control surrogate models during a run. In particular, they adaptively select a surrogate model suitable for problems or search situations. This selection is performed each time before use \cite{Li2022-bl}. Typically, adaptive SAEAs extract test data from the archive where evaluated solutions are stored and select one surrogate model with the minimum approximation error using the mean squared error (MSE) or root MSE (RMSE). 

The model type is the representative adaptation target, e.g., SUMO \cite{Gorissen2009-av} constructs seven surrogate models using different ML models like RBFN, Kriging, ANN, etc., then selects a surrogate model with the minimum MSE obtained by test data. Most work on adaptation of surrogate model settings is based on RBFNs. RBFBS \cite{Li2019-db} samples 11 different values of the spread parameter of RBF and changes the smoothness of the approximated fitness landscape. It constructs the same number of surrogate models and one surrogate model with the smallest RMSE is selected via leave-one-out cross-validation. Five surrogate models are created with five types of RBFs in aRBF-NFO \cite{Yu2021-nb} to select one surrogate model with the best RMSE. Unlike SAEAs with an adaptation of model type or settings, SADE-ATDSC defines four sampling criteria of training data for surrogate model construction and constructs surrogate models specialized for different decision variable areas. Again, a surrogate model with the minimum RMSE is selected. 
Another direction of adaptation target is the type of EA. For example, SAHO \cite{Pan2021-jj} and SAF-TD \cite{Lin2024-sp} change the EA type from DE and Teaching-Learning-based Optimization \cite{Rao2013-ri} when the performance of used EA stagnates. SAF-TD also adapts some hyperparameters of the selected EA type. Similarlly, AKAEA \cite{Lu2025-yy} adaptively selects the member of subpopulations and their search strategies based on objective values and diversity of solutions. HSADE-CS \cite{Yu2025-wr} adapts the DE hyperparameters by extending the adaptation algorithm of L-SHADE \cite{Tanabe2014-el}. AutoSAPSO \cite{Dai2025-jv} adapts both the model type and the EA type along with its hyperparameters from RBFN or Kriging and five Particle Swarm Optimization (PSO) \cite{Kennedy1995-qo} variants, respectively.

Recently, a few adaptive SAEAs do not directly use the approximation error for adaptation. In DDEA-MESS \cite{Yu2022-iw}, selection probabilities are scheduled in advance based on experimentally observed findings so that the adaptation target varies with the number of FEs. ESA \cite{Zhen2023-ta} and AutoSAEA \cite{Xie2024-qk} use reinforcement learning methods such as Q-learning and the multi-armed bandit problem, respectively. 
For the adaptation target, DDEA-MESS and ESA predefine multiple sampling strategies for FEs, i.e., infill criteria. AutoSAEA prepares eight combinations of four model types and two infill criteria. CoE-SAEA \cite{Xie2025-qo} adapts the same targets as AutoSAEA but uses a large language model and a roulette selection.
In MOPs, AIEA \cite{Wei2025-jf} adapts infill criterion named influence degree, by calculating the distance relationship between the candidate solutions and predefined reference points in objective space. For expensive multi-task optimizations, $(\theta_{l}, \theta_{u})$-PMTO \cite{Wei2025-wb} extends SAEAs by modeling continuous task variations, enabling adaptive knowledge transfer across various tasks.

\subsection{Ensemble SAEAs}

Ensemble surrogate models designed like the weighted average of outputs of different surrogate models have strengths in prediction accuracy and robustness \cite{Goel2007-ya,Jin2011-fd}. This is effective especially when the characteristics of real engineering problems are unknown \cite{Li2019-ff}. 
ESAMA, GS-SOMA, SPOT2 \cite{Bartz-Beielstein2017-ok}, CAL-SAPSO \cite{Wang2017-ku}, and SAEA-UGC \cite{Liu2021-en} ensemble two or three different model types, such as RBFN, Kriging, and PR. This provides SAEAs a synergy effect of model types of different natures. For example, Kriging and PR are good at approximating objective functions in low-dimensional cases while RBFN is in high-dimensional cases \cite{Jin2001-se}. Kriging and PR suit a wide and narrow solution spaces, respectively \cite{Jin2001-se}.

The other directions are ensembles of surrogate models with different ML model settings, training datasets, or infill criteria. Jin \textit{et al.} \cite{Jin2004-nh} constructed a weighted average ensemble surrogate model of local ANNs to improve the approximation accuracy in local regions. The ANN structures and hyperparameters were optimized using a genetic algorithm (GA) \cite{Holland1992-xj}. For the same purpose, LES-CDE \cite{Jin2015-vy} uses an ensemble of local extreme learning machines, 
where each extreme learning machine is constructed with different local training datasets. In EAPSO and DS-SAEA \cite{Ma2025-lq}, two and three RBFNs for the ensemble differ their RBFs from each other, respectively, to utilize different characteristics. SADE-DSDS \cite{Liu2025-mc} introduces a novel infill criterion named neighborhood centrality in addition to the predicted fitness value by surrogate models to reduce the negative effect of the poor prediction accuracy. Then, it selects nondominated solutions of these two criteria for FEs. CESAEA \cite{Guo2025-nq} and CCoEA-ASAA \cite{Wei2026-zw} are for EOPs in a distributed data environment and ensembles surrogate models of clients. CESAEA employs classification models to improve robustness in a distributed environment. CCoEA-ASAA repeatedly constructs RBFNs using the bootstrap sampling during waiting time for the completion in all clients, and then aggregates them in an original manner.

In research for MOPs, SIDSAEA \cite{Zhang2025-fh} constructs two ensemble models of multiple Kriging models: one approximates the fitness values and the other does the newly proposed model management metric to improve both optimization performance and model quality. SMEA-PF \cite{Li2024-jm} ensembles surrogate models of different objectives to approximate the Pareto front.

When SAEAs conduct an ensemble, the uncertainty is available by considering the difference in outputs of surrogate models \cite{Jin2011-fd}. Consequently, CAL-SAPSO and SAEA-UGC select two solutions with the best-approximated fitness value and the largest uncertainty among candidate solutions. EAPSO uses LCB to prescreen solutions generated by PSO.

The ensemble SAEAs introduced above, except SPOT2, SADE-DSDS, and CESAEA, employ the bagging strategy to realize an ensemble. As other examples, XGBEA for MOPs uses the boosting strategy, and SPOT2 for SOPs adopts the stacking strategy by utilizing a meta-model that inputs the outputs of linear model, random forest, and Kriging.

\subsection{Adaptive Ensemble SAEAs}

Some SAEAs benefit from both adaptation and ensemble approaches. 
ASMEA first selects the surrogate models with smaller RMSEs among ones constructed with five model types and considers them elite. Next, they select one surrogate model with the smallest RMSE from the elite surrogate models and their ensemble model. HESNFO is similar to aRBF-NFO in the adaptation of five RBFs, however its selection method slightly differs from that of aRBF-NFO. After calculating the RMSE on the test data for the five surrogate models, HESNFO selects the top two surrogate models and ensembles them by averaging their outputs of the predictions. 
DSP-SAEA \cite{Liu2023-ml} adapts model types in local search with PSO. Three surrogate models with RBFN, Kriging, and PR and their ensemble model are constructed in the trust region \cite{Ong2003-sa}, and one with the minimum RMSE is used. PS-SAEA \cite{Wang2025-jb} is an SAEA for offline EOPs, wherein a certain number of initial data is given and no additional FEs can be conducted during the optimization. It constructs 2,000 RBFNs using the bootstrap sampling from the initial data at the beginning of the optimization. Then, it ensembles some RBFNs randomly selected using the probability calculated from the RMSEs.

In the literature on SAEAs for MOPs, ExTrEMO \cite{Liu2025-qz} utilizes knowledge from similar source tasks and aggregates it by transfer Gaussian process surrogate models. The source tasks are selected adaptively with relevance to the target task. MLPSGP-SAEA \cite{Pan2025-md} ensembles the LCB and Expected Improvement (EI) infill criteria by weighting with the sparsity and diversity of solutions, respectively.

\subsection{Position of \ours}

As shown above, the adaptation targets and ensemble elements in existing SAEAs are widely designed. However, the surrogate model selection methods are highly dependent on the approximation error and only one surrogate model is used in SAEAs with adaptation. Conversely, in ensemble SAEAs, the degree of smoothness of the approximated fitness landscape is not carefully controlled. As shown in the ``Number of Seeds'' column, a limited number of variant seeds are employed in the related work. This is because of the manual predefinition of adaptation candidates and ensemble elements. This results in an inflexible adaptation or control of the degree of smoothness of the approximated fitness landscapes. The suitability of these variant seeds is not verified upfront for the given problems. 

Therefore, we design the proposed \our, categorized in adaptive ensemble SAEAs, as shown in the bottom row of TABLE~\ref{tab:rtw}. \ours optimizes the RBFN structure using NSGA-II to control the degree of smoothness of the approximated fitness landscapes flexibly for the first time in the SAEA research lineage. This provides a high degree of freedom in tuning the degree of smoothness and a much larger number of surrogate models ($110$ in this work) than existing SAEAs. This number is equal to the population size of NSGA-II times the number of generations plus the number of initial solutions of NSGA-II. Different from existing adaptation approaches, simultaneous minimization of the approximation error and model complexity hedges the risk of over-reliance on approximation error and validates the suitability of the surrogate models elaborately.

\section{Preliminary} \label{sec:tch}

This section introduces DE and NSGA-II. Then we describe the structure of the RBFN used in this work.

\subsection{DE: Differential Evolution} \label{sec:tch-de}

DE is an evolutionary algorithm for real-parameter bound-constrained SOPs, where the objective function is $f:\mathbb{R}^{D}\to \mathbb{R}$, $D$ is the problem dimension, and a solution is $\bm{x} =[x_{1}, \dots, x_{D}]^\mathsf{T} \in \mathbb{R}^{D}$. The details of the algorithm of DE including the employed bound-constraint handling technique are described in Section S-I-A of the Supplementary material.

\subsection{NSGA-II: Nondominated Sorting Genetic Algorithm II} \label{sec:tch-nsgaii}

NSGA-II is a multi-objective evolutionary algorithm (MOEA) for solving MOPs with a small number of objectives. Here, we assume real-parameter bound-constrained MOPs.
The detailed algorithm of NSGA-II including the employed bound-constraint handling technique is given in Section S-I-B of the Supplementary material.

\subsection{RBFN: Radial Basis Function Network} \label{sec:tch-rbfn}

An RBFN is a three-layer feed-forward ANN. The input layer has the same number of nodes as dimensions $D$ of the training data and conveys the data to the hidden layer. The hidden layer calculates the RBF values. The Gaussian function is a representative RBF, and this RBF of the $j$-th node, ${\phi}_{j}(\bm{x})$, can be expressed as:
\begin{equation}
    {\phi}_{j}(\bm{x})=\exp\left(-\left\| \bm{x} - \bm{c}_{j} \right\|^2 / 2{\sigma}^{2}\right),
    \label{eq:rbf}
\end{equation}
where $\bm{c}_{j}$ is the center point and $\sigma$ is the spread parameter. The output layer yields the linear combination of the hidden layer nodes as:
\begin{equation}
	\hat{f}(\bm{x}) = \sum^{n}_{j=1} {w}_{j} {\phi}_{j}\left(\bm{x}\right),
	\label{eq:rbf-output}
\end{equation}
where $\hat{f}(\bm{x})$ is the output as a prediction of $f(\bm{x})$, ${w}_{j}$ is the $j$-th element of the weight coefficient vector $\bm{w} = [w_{1}, \dots, w_{n}]^\mathsf{T}$, and $n$ is the number of hidden layer nodes.

Appropriate settings of the four parameters $n$, $\bm{c}_{j}$, $\sigma$, and $\bm{w}$ are necessary to improve the performance of RBFN. The weight $\bm{w}$ is usually obtained by solving $\bm{w}=\bm{\Phi}^{-1} \bm{F}$, where $\bm{\Phi}$ is an $N \times n$ matrix whose elements are ${\phi}_{j}(\bm{x}_{i})$, $N$ is the size of the training data, and $\bm{F}=[{f}(\bm{x}_{1}), \dots,{f}(\bm{x}_{i}), \dots, {f}(\bm{x}_{N})]^\mathsf{T}$ is a vector of objective function values of the training data. Conversely, the values of $n$ and $\sigma$ affect the smoothness of the approximated fitness landscape, and the choice of $\bm{c}_{j}$ determines the region of strong response in the input space \cite{Park1991-et}. There is abundant literature on how to set the values of $n$, $\bm{c}_{j}$, and $\sigma$ \cite{Murata1999-be}. 

\section{\our} \label{sec:our}

This section describes the concept of designing \ours and elaborates on the algorithm details.

\subsection{Concept} \label{sec:our-concept}

\begin{figure}[t]
    \includegraphics[width = 0.49\textwidth]{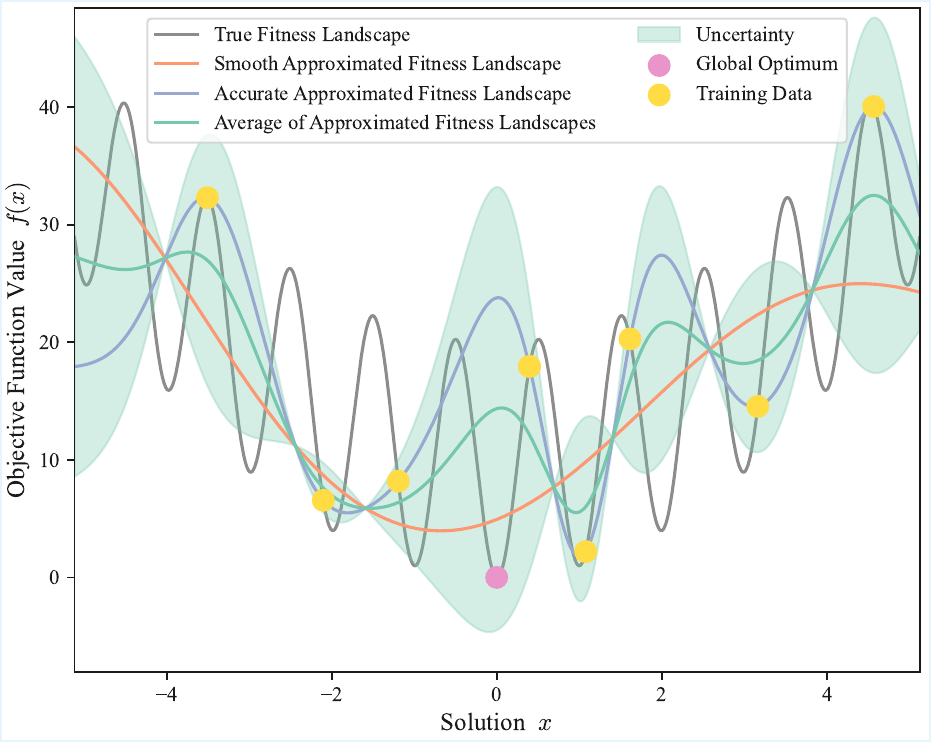}  
    \vspace{-5mm}
    \caption{A demonstration of the effectiveness of the ensemble surrogate model with an emulation of uncertainty on the one-dimensional Rastrigin problem. Two surrogate models are constructed from training data (yellow points) to approximate the true fitness landscape (black line). They are trained to derive different approximated fitness landscapes; a smooth one (orange line) and an accurate one (blue line). They consist of an ensemble model with an average landscape (green line) and uncertainty (green shade). This ensemble model is expected to help \ours find the global optimum (pink point) efficiently.}
    \label{fig:ensemble_landscape}
\end{figure}

\ours is designed to automatically prepare surrogate models with different degrees of smoothness of the approximated fitness landscapes while maintaining their prediction accuracy. Fig.~\ref{fig:ensemble_landscape} demonstrates the effectiveness of the ensemble model with an emulation of uncertainty on the one-dimensional Rastrigin problem, whose fitness landscape and global optimum are drawn with the black line and pink point, respectively. For example, we show two different surrogate models, which are prepared by \our, having relatively smooth and accurate approximated fitness landscapes with the orange and blue lines, respectively. In the figure, they were actually obtained from the training data shown in the yellow points. When they are ensembled by averaging their outputs, the fitness landscape of the ensemble model, drawn with the green line, combines these two characteristics and can effectively guide solutions to the global optimum as follows. The smooth approximated fitness landscape tries to correct the global shape as well as the place of the deepest valley of the accurate approximated fitness landscape while mitigating the effects of surrogate models that provide abnormal predictions. This element of the ensemble moves solutions to the direction to the region where the global optimum exists. Conversely, the accurate approximated fitness landscape tries to chase the detailed shape of the true fitness landscape, which helps to find the exact point of the global optimum in local regions. The differences between green and orange or blue, i.e., the model disagreement, can be utilized as the uncertainty, while avoiding the excessive computational cost and errors even in high-dimensional cases \cite{Guo2019-at}, compared to the formal probabilistic modeling in Kriging. Unlike Kriging, the convergence of this emulated uncertainty is not theoretically guaranteed as the number of samples increases. However, similar to Kriging, leveraging the emulated uncertainty effectively enhances prediction accuracy as well as the ability to recommend high-quality solutions. Its validity is empirically demonstrated in the literature of Deep Ensemble \cite{Lakshminarayanan2016-dr}. By adding or subtracting a constant (e.g., two) multiple of its value from the average, uncertainty can be expressed as shown in the green shade in the figure. Utilizing it has the advantage of correcting when the predictions of surrogate models are conflicting; when multiple surrogate models are generated, different approximated fitness landscapes often predict opposite peaks and valleys within the same interval, as shown in the figure. Under the LCB infill criterion, the solution with the smallest lower edge of the green shade is selected, which likely helps \ours obtain a solution close to the global optimum. In this way, the proposed mechanisms are expected to conduct an efficient optimization.

Following these discussions, to realize the ensemble of accurate surrogate models with a variety of degrees of smoothness of the approximated fitness landscapes, \ours is designed to obtain a set of Pareto-optimal surrogate models by minimizing the approximation error and model complexity with NSGA-II. Specifically, \ours adapts the number of hidden layer nodes $n$ and the spread parameter $\sigma$ of the RBFN, which decides the degree of smoothness of the approximated fitness landscape. Usually, $n$ and $\bm{c}_{j}$ are set to the training data size and each training data point, respectively, in most SAEAs \cite{Nishihara2026-qi, Liu2023-ml, Yu2022-wl, Xie2024-qk, Yu2021-nb, Nishihara2022-nr, Nishihara2023-wr, Yu2020-oz, Yu2022-iw, Zhen2023-ta}. However, an RBFN with a large number of hidden layer nodes has high expressive power, but sometimes degrades generalization performance \cite{Murata1999-be}. Additionally, over fitting can occur when the training data are very close to each other \cite{Wang2017-ku}. In \our, while $\sigma$ and $n$ are evolutionarily optimized using NSGA-II, the $n$ center points $\{\bm{c}_{j}\}^{n}_{j=1}$ are set to cluster centers obtained by the k-means method to avoid excessive concentration.

\subsection{Mechanism} \label{sec:our-alg}

\begin{figure}[t]
    \includegraphics[width = 0.5\textwidth]{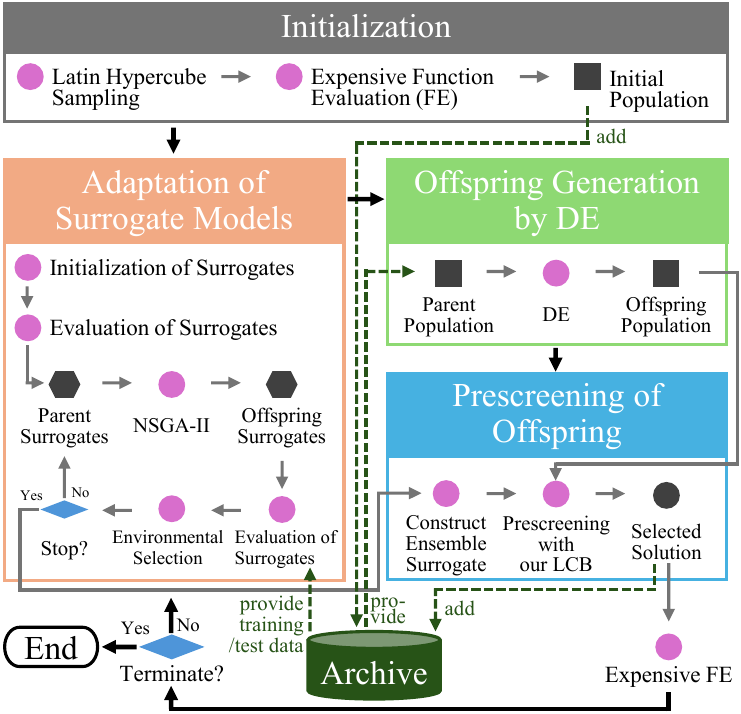}
    \caption{A diagram of the proposed \our, where the solid arrows stand for the flows of the algorithm and the dashed arrows stand for the data flows.}
    \label{fig:diagram}
\end{figure}

Algorithm~\ref{alg:our} shows the pseudo-code of the \ours procedures and Fig.~\ref{fig:diagram} illustrates a diagram of \our. First, \ours initializes its population $\mathcal{P}=\{\bm{x}_i\}^{N}_{i=1}$ with Latin Hypercube Sampling (LHS). Next, all solutions in $\mathcal{P}$ are evaluated and added to an archive $\mathcal{A}=\{\left(\bm{x}_{i}, f(\bm{x}_{i})\right)\}_{i=1}^{N}$, where all evaluated solutions are stored. In the successive main loop, \ours repeats three phases: adaptation of surrogate models, offspring generation by DE, and prescreening of offspring, at each generation until the termination criteria are met.

\subsubsection{Adaptation of Surrogate Models}

\ours constructs surrogate models that promote search in possible regions to improve the fitness value. First, the top $N_{\mathit{data}}$ solutions are extracted from $\mathcal{A}$ and form $\mathcal{D}$. Then, $\mathcal{D}$ is divided into training data $\mathcal{D}_{\mathit{train}}$ and test data $\mathcal{D}_{\mathit{test}}$ by the ratio of $(1-\delta) : \delta$, where $\delta \in [0,1]$. Successively, \ours prepares for the optimization of surrogate models conducted with NSGA-II. 

\ours solves the following MOP to adapt surrogate models:
\begin{align}
    \hspace{-3mm}
	\bm{F}(\bm{x}_{M}, \mathcal{D}_{\mathit{test}}) & =\hspace{-2pt} \{{f}_{1}(\bm{x}_{M},\hspace{-1pt} \mathcal{D}_{\mathit{test}}), {f}_{2}(\bm{x}_{M})\} \hspace{-2pt}=\hspace{-2pt} \{\hspace{-1pt}\mathit{RMSE}, n\}, \label{eq:MOP}\\
	\bm{x}_{M} & = [{x}_{M,1}, {x}_{M,2}]^\mathsf{T} = [n, \sigma]^\mathsf{T}, \label{eq:MOP_x}\\
    \mbox{s.t. } {x}_{M,1} & \in [|\mathcal{D}_{\mathit{train}}|/2, |\mathcal{D}_{\mathit{train}}|], \label{eq:MOP_x1}\\
    {x}_{M,2} &\in [\min \mathit{dist}(\mathcal{D}_{\mathit{train}}), \max \mathit{dist}(\mathcal{D}_{\mathit{train}})], \label{eq:MOP_x2}
\end{align}
where $\mathit{dist}(\mathcal{D}_{\mathit{train}})$ is a set of distances between every pair of the training data points. The evaluation in this MOP equals measuring approximation error ${f}_{1}(\bm{x}_{M}, \mathcal{D}_{\mathit{test}})$ and model complexity ${f}_{2}(\bm{x}_{M})$ of RBFNs (Eq.~(\ref{eq:MOP})). We use $\mathit{RMSE}$ calculated with the test data $\mathcal{D}_{\mathit{test}}$ as the most representative approximation error metric in adaptive SAEAs \cite{Liu2023-ml, Yu2022-wl, Yu2021-nb, Nishihara2022-nr, Yu2020-oz}. Considering the structure of RBFN, the complexity is set to the number of hidden layer nodes $n$. As mentioned before, two decision variables are the number of hidden layer nodes ${x}_{M,1}=n$ and the spread parameter ${x}_{M,2}=\sigma$ (Eq.~(\ref{eq:MOP_x})). The search range of ${x}_{M,1}$ is set as Eq.~(\ref{eq:MOP_x1}). Specifically, the minimum value is set to half the training data size to ensure accuracy. The maximum value is set to the training data size as the possible number of center points. The spread parameter $\sigma$ stands for the radius from the center point. A new point in the radius is affected by the node. Therefore, the search range of ${x}_{M,2}$ is designed to be from the minimum to maximum distances as Eq.~(\ref{eq:MOP_x2}) so that the minimum of two points and the maximum of all points affect a new point.

Then, the initial population $\mathcal{P}_{M}=\{\bm{x}_{M,i}\}_{i=1}^{{N}_{M}}$ for NSGA-II is generated with the LHS and evaluated. Note that ${x}_{M,1}$ is treated as a real value during NSGA-II optimization, but is rounded to the nearest integer during evaluation. At the end of the NSGA-II initialization, the solutions in the population are ranked by the NDCD to prepare for the tournament selection. NSGA-II optimizes RBFN over ${\omega}_{\max}$ generations by applying the tournament selection, SBX, and PM shown in Section~\ref{sec:tch-nsgaii}. The Pareto set $\mathcal{PS}$ obtained in the last generation is saved as a set of Pareto-optimal surrogate models.

\subsubsection{Offspring Generation by DE}

In this phase, the top $N$ solutions are extracted from $\mathcal{A}$ and form the population $\mathcal{P}$ of DE. Subsequently, a set of offspring $\mathcal{Q}_{M} = \{\bm{u}_{i}\}_{i=1}^{N}$ is generated using the \textit{best/1} mutation and the \textit{binomial} crossover strategy introduced in Section~\ref{sec:tch-de}. Since EOPs requires exploitation tendency for evolutionary operators \cite{Tong2021-ay,Nishihara2024-ol}, we used \textit{best/1} having a better exploitation ability \cite{Storn1997-ny,Nishihara2024-ol}. Note that FEs and solution updates are performed in the next phase.

\subsubsection{Prescreening of Offspring}

This phase prescreens the solution among the offspring generated by DE for actual FE utilizing surrogate models in the aforementioned $\mathcal{PS}$. We employ the prescreening strategy instead of the iterative search strategy to maintain the diversity of the evaluated solutions for constructing promising ensemble surrogate models. The performance comparison between the prescreening and iterative search strategies on the \ours framework is described in Section S-VII of the Supplementary material. The ensemble model construction and prescreening are performed as follows.

\paragraph{Construction of Ensemble Model}
First, an ensemble model of all surrogate models in $\mathcal{PS}$ is constructed by averaging all outputs and calculating the uncertainty as:
\begin{align}
	\bar{f}(\bm{x}) & = \frac{\sum\limits_{i=1}^{|\mathcal{PS}|} \hat{f}_{i}(\bm{x})}{|\mathcal{PS}|}, \label{eq:mean}\\
	\hat{s}\left(\hat{f}(\bm{x})\right) & = \sqrt{\frac{\sum\limits_{i=1}^{|\mathcal{PS}|} \left(\hat{f}_{i}(\bm{x})-\bar{f}(\bm{x})\right)^2}{|\mathcal{PS}|-1}}, \label{eq:sd}
\end{align}
where $\hat{f}_{i}(\bm{x})$ is an output of a surrogate model in $\mathcal{PS}$, $\bar{f}(\bm{x})$ is the mean of all outputs, and $\hat{s}\left(\hat{f}(\bm{x})\right)$ is the unbiased estimation of the standard deviation.

\paragraph{Calculation of LCB for Offspring}
We use LCB for the infill criterion in prescreening, defined as:
\begin{equation}
	\mathit{LCB}(\bm{x}) = \bar{f}(\bm{x}) - \alpha \ \hat{s}\left(\hat{f}(\bm{x})\right), \label{eq:LCB}
\end{equation}
where $\alpha$ is the coefficient as a hyperparameter. \ours calculates the LCB values of $\forall \bm{x} \in \mathcal{Q}_{M}$.

\paragraph{Prescreening and Evaluation}
An offspring deriving the smallest LCB is selected, which is marked as $\bm{u}^{*}$. Finally, $\bm{u}^{*}$ is evaluated with the objective function $f$ and added to $\mathcal{A}$.

We discuss the time complexity of \ours in Section S-II of the Supplementary material, with comparison to normal, adaptive, and ensemble SAEAs. The results show that \ours has a larger complexity than normal SAEAs, while the settings of $N_{M}$ and ${\omega}_{\max}$ affect the degree of complexity of \ours over that of adaptive/ensemble SAEAs.

\begin{algorithm}[t]
    \caption{\ours} \label{alg:our}
    \small
	\begin{algorithmic}[1]
        \REQUIRE the maximum number of function evaluations $\mathit{FE}_{\max}$, population size $N$, scaling factor for DE $F$, crossover rate for DE $\mathit{CR}$, data size for surrogate modeling $N_{\mathit{data}}$, test-data rate $\delta$, population size for NSGA-II $N_M$, the maximum number of generations for NSGA-II ${\omega}_{\max}$, distribution index of crossover for NSGA-II ${\eta}_{c}$, distribution index of mutation for NSGA-II ${\eta}_{m}$, and coefficient of LCB $\alpha$
		\STATE Initialize $\hspace{-2pt}\mathcal{P}\hspace{-2pt}=\hspace{-2pt}\{\bm{x}_{i}\}_{i=1}^{N}\hspace{-2pt}$ by LHS and Evaluate $\forall \bm{x} \hspace{-2pt}\in\hspace{-2pt} \mathcal{P}$ with true $f$
		\STATE $\mathcal{A} = \{\left(\bm{x}_{i}, f(\bm{x}_{i})\right)\}_{i=1}^{N}$, $\mathit{FE}=N$
		\WHILE{$\mathit{FE}<\mathit{FE}_{\max}$}
            \REMARK{\textbf{Phase 1}: Adaptation of Surrogate Models}
            \STATE $\mathcal{D} \gets$ Get top $N_{\mathit{data}}$ data from $\mathcal{A}$ \label{alg-l:adaptation-begin}
            \STATE Divide $\mathcal{D}$ into training/test datasets in the ratio $(1-\delta):\delta$
            \STATE Set the NSGA-II boundary \hfill | Eqs.~(\ref{eq:MOP_x1})\&(\ref{eq:MOP_x2}) \label{alg-l:MOEA-begin}
            \STATE Initialize $\mathcal{P}_{M}=\{\bm{x}_{M,i}\}_{i=1}^{{N}_{M}}$ by LHS
            \STATE Construct RBFNs $\forall \bm{x}_{M} \in \mathcal{P}_{M}$ and Evaluate \hfill | Eq.~(\ref{eq:MOP})
            \STATE Perform NDCD on $\mathcal{P}_{M}$
            \FOR{$\omega=1$ \TO ${\omega}_{\max}$}
                \STATE $\mathcal{Q}_{M} \gets$ Generate ${N}_{M}$ offspring from $\mathcal{P}_{M}$ with ${\eta}_{c}$ \& ${\eta}_{m}$
                    \\ \hfill | Eqs.~(S-3)\&(S-4)
                \STATE Construct RBFNs $\forall \bm{x}_{M} \in \mathcal{Q}_{M}$ and Evaluate \hfill | Eq.~(\ref{eq:MOP})
                \STATE Perform NDCD on $\mathcal{R}_{M}=\mathcal{P}_{M}\cup\mathcal{Q}_{M}$
                \STATE $\mathcal{P}_{M} \gets$ Select top ${N}_{M}$ solutions from $\mathcal{R}_{M}$
            \ENDFOR
            \STATE $\mathcal{PS} \gets$ Get a Pareto set from $\mathcal{P}_{M}$ \label{alg-l:MOEA-end}

            \REMARK{\textbf{Phase 2}: Offspring Generation by DE}
            \STATE $\mathcal{P} \gets$ Get top $N$ solutions from $\mathcal{A}$
            \STATE $\mathcal{Q} \gets$ Generate $N$ offspring from $\mathcal{P}$ with $F$ \& $\mathit{CR}$
                \\ \hfill | Eqs.~(S-1)\&(S-2)
            
            \REMARK{\textbf{Phase 3}: Prescreening of Offspring} 
            \STATE Construct an ensemble model of all in $\mathcal{PS}$ \hfill | Eqs.~(\ref{eq:mean})\&(\ref{eq:sd}) \label{alg-l:prescreening-begin}
            \STATE $\bm{u}^{*} \gets \argmin_{\bm{u}\in\mathcal{Q}} \mathit{LCB}(\bm{u})$ calculated with $\alpha$ \hfill | Eq.~(\ref{eq:LCB})
            \STATE $f(\bm{u}^{*}) \gets$ Evaluate $\bm{u}^{*}$ with true $f$
            \STATE $\mathcal{A} = \mathcal{A} \cup \{(\bm{u}^{*}, f(\bm{u}^{*}))\}$, $\mathit{FE}=\mathit{FE}+1$ \label{alg-l:prescreening-end}
		\ENDWHILE
        \ENSURE The best solution in $\mathcal{A}$
	\end{algorithmic}
\end{algorithm}

\section{Experimental Results} \label{sec:exp}

This section assesses the efficacy of \ours by comparing its performance with that of state-of-the-art SAEAs, primarily through the Wilcoxon signed-rank test.

\subsection{Experimental Design}

\subsubsection{Problem Settings}

This work used the benchmark function suite proposed in the IEEE CEC 2020 Special Session and Competition on Single Objective Bound Constrained Numerical Optimization \cite{Yue2019-vz}, which comprises a uni-modal function F1, multi-modal functions F2-F4, hybrid functions F5-F7, and composition functions F8-F10. We set the problem dimension to $D \in \{30, 50, 100\}$ to evaluate whether \ours has a scalability of the performance to the increase of $D$. Hence, we conducted $30$ cases ($10 \times 3$) in total. Following the regulation of competition \cite{Yue2019-vz}, the search range (bound constraint) was set to $[-100, 100]^D$ for all functions. All experiments were conducted with the Intel Core i7-10700 (2.9 GHz) CPU and 16 GB RAM and the PlatEMO software \cite{Tian2017-bc}.

\subsubsection{Compared Algorithms}

We compared \ours to GL-SADE \cite{Wang2023-dc}, IDRCEA \cite{Li2023-ns}, TS-DDEO \cite{Zhen2023-jy}, and MHS-QPSO \cite{Li2025-io} as normal modern SAEAs for high-dimensional problems, SADE-ATDSC \cite{Nishihara2022-nr} and AutoSAEA \cite{Xie2024-qk} as adaptive SAEAs, CAL-SAPSO \cite{Wang2017-ku} and EAPSO \cite{Li2019-ff} as ensemble SAEAs, and HESNFO \cite{Yu2022-wl} and DSP-SAEA \cite{Liu2023-ml} as adaptive ensemble SAEAs. For SAEAs with adaptation mechanisms, because the adaptation methods differ from each other, their impact on performance can be compared. In particular, SADE-ATDSC, HESNFO, and DSP-SAEA select surrogate models with the minimum approximation error and AutoSAEA selects one by reinforcement learning. Thus, we can verify the effectiveness of using both accurate and less complex surrogate models in \our. For SAEAs with ensemble mechanisms, surrogate models to be aggregated are evolutionarily optimized in \ours unlike CAL-SAPSO, EAPSO, HESNFO, and DSP-SAEA, so the impact of flexible automatic design of ensemble elements can be observed. All the compared algorithms use the RBFN. Furthermore, EAPSO uses the LCB, which is the same infill criterion to that of \our.
Hyperparameter settings followed the original papers and we show those of compared algorithms and \ours in TABLE S-II in the Supplementary material.

\subsubsection{Evaluation Criteria}

We set the maximum number of FEs to $1,000$, considering EOPs that allow for expensive evaluations up to approximately $1,000$ times, such as the Blended-Wing-Body underwater vehicles optimization \cite{Dong2021-cs} and the oil reservoir production optimization problem \cite{Xie2024-qk,Li2023-ns}, and report the performance with the average fitness values of $31$ independent trials. We also applied the Wilcoxon signed-rank test with a significance level of $0.05$ to detect statistical significance. These statistical results are reported with ``$+$,'' ``$-$,'' or ``$\sim$,'' which means the compared algorithm significantly outperformed \our, significantly underperformed \our, or we cannot find a significant difference, respectively. 

\subsection{Result}

TABLE~\ref{tab:data_summary} is the summary of results obtained at $1,000$ FEs for $D \in \{30,50,100\}$, while the detailed results of the average fitness values and standard deviations (in parentheses) are provided in TABLE S-VII of the Supplementary material. The best and worst values among the eleven algorithms are highlighted in green bold and pink italics, respectively.

For $D=30$, \ours was competitive with AutoSAEA and DSP-SAEA because the differences in the numbers of ``$-$'' and ``$+$'' indicating the inferiority and superiority of compared algorithms, respectively, were one and two, respectively. However, \ours outperformed GL-SADE, IDRCEA, TS-DDEO, MHS-QPSO, SADE-ATDSC, CAL-SAPSO, EAPSO, and HESNFO in most problems. When $D$ increased to $50$, the performance of \ours improved much better than that of DSP-SAEA because \ours statistically outperformed DSP-SAEA on all the $10$ problem instances. Similarly, \ours was also statistically better than AutoSAEA in the half problem instances of the benchmark suite. This tendency was even more pronounced at $D = 100$. The statistical results against AutoSAEA and DSP-SAEA were $+\hspace{-1pt}/\hspace{-1pt}-\hspace{-1pt}/\hspace{-1pt}\sim=1/7/2$ and $1/8/1$, respectively, which demonstrates the better performance scalability of \ours to the increase of $D$ than that of AutoSAEA and DSP-SAEA. Moreover, \ours derived five to ten ``$-$'' against each compared algorithm and one or zero ``$+$,'' indicating the superiority of \ours in higher-dimensional problems. On the other hand, TS-DDEO became effective when $D$ increased to $100$ as it statistically outperformed and underperformed \ours on four and five problems, respectively. However, the performance of TS-DDEO was unstable in high-dimensional cases since it derived both the best (green bold) and worst (pink italic) performance among all algorithms while \ours achieved the best performance but did not obtain the worst. These results demonstrate that the performance of \ours was highly scalable to the increase of problem dimension $D$.

The overall trend is as follows. As shown in TABLE~\ref{tab:data_summary}, in all pairs of \ours and each compared algorithm, the number of ``$-$'' outnumbered the number of ``$+$'' with a minimum and maximum difference of $10$ and $30$, respectively. The total of $300$ comparisons across the ten compared algorithms and the three types of dimension showed that $+\hspace{-1pt}/\hspace{-1pt}-\hspace{-1pt}/\hspace{-1pt}\sim=24/251/25$. Thus, our algorithm derived statistically superior performances on many problem instances. Besides, \ours obtained the best average rank.

One drawback is that \ours was not specialized for different problems under lower-dimensional cases. \ours derived fewer times of the best performance (highlighted in green bold) than AutoSAEA and DSP-SAEA at $D=30$. \our, AutoSAEA, and DSP-SAEA were the best on one, five, and three problems, respectively. This indicates that \ours should consider other model types like Kriging in lower-dimensional cases. This is due to the fact that AutoSAEA and DSP-SAEA use multiple model types and contribute to better overall performance.
However, this drawback improved as the problem dimension increased and \ours derived a larger number of the best performance than AutoSAEA and DSP-SAEA at $D=100$. Notably, \ours was the best for every category of problems, i.e., basic (F2), hybrid (F6), and composition (F8) functions, indicating the good generalization performance of \our. RBFN was useful and the automatic design of ensemble surrogate model of \ours was effective in higher-dimensional cases. \ours derived better statistical test results and the best rank than any compared algorithm for all $D$, which demonstrates the high robustness and superior optimization performance of \ours for every $D$. We further discussed the results from the point of view of the difference in adaptation/ensemble method in Section S-VI-B of the Supplementary material.

To evaluate the convergence performance of \our, we show the results of the Wilcoxon signed-rank tests at $200$, $400$, $600$, and $800$ FEs in addition to $1,000$ FEs (baseline) for $D\in\{30, 50, 100\}$ in TABLE~\ref{tab:sigs}. We also display the convergence curves of the performance in Fig. S-I of the Supplementary material.
In the table, although \ours underperformed or was competitive to TS-DDEO, MHS-QPSO, and AutoSAEA for some cases in $D = 50$-$100$ and $200$-$800$ FEs, the number of ``$-$'' exceeded that of ``$+$'' in the other cases. For the entire results, the differences between the numbers of ``$-$'' and ``$+$'' remained stable or increased in most cases as the number of FEs increased. 
The figure also demonstrates the good convergence ability of \our, which can be confirmed with the competitive or superior convergence speed of \ours to the compared algorithms on many problem instances, especially after $500$ FEs. Although \ours did not derive the best convergence speed under an extremely strict budget like $200$ FEs, the statistical test demonstrated the outperforming performance of \ours at $200$ FEs over the compared algorithms except for MHS-QPSO.
Therefore, we confirmed that the performance of \ours scales not only with $D$ but also with the number of FEs. Specifically, \ours works well even when the data size is limited or FEs are allowed to $1,000$ times.

Lastly, TABLE~\ref{tab:runtime} summarizes the average computational time required for completing optimizations with $1,000$ FEs obtained by eleven algorithms. Each value is averaged for $31$ trials and all problems in each dimension. While SADE-ATDSC and HESNFO derived a relatively short computational time than the others, CAL-SAPSO, DSP-SAEA, and \ours took relatively longer. \ours spent a long computational time because of the evolutionary adaptation of surrogate models in the automatic design of ensemble surrogate models. Moreover, CAL-SAPSO and DSP-SAEA required much longer computational time than \ours because the construction and prediction of PR and Kriging take a lot, especially in medium to high-dimensional cases. However, the computational time of \ours is still acceptable in EOPs as it required approximately $9.57$ seconds per one FE at maximum while one FE in real-world EOPs often requires more than one hour \cite{Le2013-gk}. Therefore, the computational time of \ours can be seen as trivial compared to expensive FEs in real-world applications.

\begin{table*}[t]
    \centering
    \caption{The summary of results at $1,000$ fitness evaluations for $D \in \{30,50,100\}$ on the CEC2020 Benchmark Suite.}
    \vspace{-2mm}
    \label{tab:data_summary}
    \centering
    \scalebox{0.9}{
        \begin{tabular}{c|cccc|cc|cc|ccc}\hline\hline
            			& \multicolumn{4}{c|}{Normal SAEAs}                                                                                                                 & \multicolumn{2}{c|}{Adaptive SAEAs}                                   & \multicolumn{2}{c|}{Ensemble SAEAs}                               & \multicolumn{3}{c}{Adaptive Ensemble SAEAs}                                                       \\
            			& GL-SADE                           & IDRCEA                            & TS-DDEO                               & MHS-QPSO                          & SADE-ATDSC                        & AutoSAEA                          & CAL-SAPSO                         & EAPSO                         & HESNFO                            & DSP-SAEA                          & \ours                     \\\hline
            $+/-/\sim$	&	    	2/28/0		            &	    	0/30/0		            &	   	7/20/3		                    &	   	0/27/3		                &		5/20/5		                &		5/15/10		                &		1/29/0		                &		0/29/1		            &		0/30/0		                &		4/23/3		                &	    	-	            \\
            Ave. Rank	&	       	6.567 		            &	       	9.700 		            &	   	5.133 		                    &	   	6.233 		                &	   	3.600 	   	                &	   	3.933 	   	                &	   	6.900 	   	                &	   	7.200 	   	            &	   \worst	10.467 	   	        &	   	3.967 	   	                &	   \win	2.300    	    \\ \hline
        \end{tabular}
    }
\end{table*}

\begin{table}[t]
    \renewcommand{\arraystretch}{1.1}
    \centering
    \caption{Significant differences regarding findings for ``$+\hspace{-1pt}/\hspace{-1pt}-\hspace{-1pt}/\hspace{-1pt}\sim$'' in comparison between \ours and state-of-the-art SAEAs.}
    \vspace{-2mm}
    {\tabcolsep = 1mm
    \label{tab:sigs}
    \begin{minipage}[t]{0.495\textwidth}
        \centering
        \scalebox{0.825}{
            \begin{tabular}{c|c|cccccccccc}\hline\hline
                $D$ & FEs       & \hspace{-2mm}\begin{tabular}{c} vs \\[-1mm] GL- \\[-1mm] SADE \end{tabular}\hspace{-2mm} & \hspace{-2mm}\begin{tabular}{c} vs \\[-1mm] IDR \\[-1mm] CEA \end{tabular}\hspace{-2mm} & \hspace{-2mm}\begin{tabular}{c} vs \\[-1mm] TS- \\[-1mm] DDEO \end{tabular}\hspace{-2mm} & \hspace{-2mm}\begin{tabular}{c} vs \\[-1mm] MHS- \\[-1mm] QPSO \end{tabular}\hspace{-2mm} & \hspace{-2mm}\begin{tabular}{c} vs \\[-1mm] SADE- \\[-1mm] ATDSC \end{tabular}\hspace{-2mm} & \hspace{-2mm}\begin{tabular}{c} vs \\[-1mm] Auto \\[-1mm] SAEA \end{tabular}\hspace{-2mm} & \hspace{-2mm}\begin{tabular}{c} vs \\[-1mm] CAL- \\[-1mm] SAPSO \end{tabular}\hspace{-2mm} & \hspace{-2mm}\begin{tabular}{c} vs \\[-1mm] EAPSO \\[-1mm] \ \end{tabular}\hspace{-2mm} & \hspace{-2mm}\begin{tabular}{c} vs \\[-1mm] HES \\[-1mm] NFO \end{tabular}\hspace{-2mm} & \hspace{-2mm}\begin{tabular}{c} vs \\[-1mm] DSP- \\[-1mm] SAEA \end{tabular}\hspace{-2mm} \\\hline
                    & \;\;200  	&   1/\;\;6/3	&	0/\;\;9/1	&	1/7/2 	&	3/\;\;6/1 	&	0/3/7	&	1/2/7	&	1/\;\;7/2	&	0/\;\;9/1	&	0/\;\;9/1	&	2/\;\;4/4	\\
                    & \;\;400  	&   0/10/0	    &	0/10/0	    &	2/6/2 	&	1/\;\;7/2 	&	2/5/3	&	1/3/6	&	2/\;\;7/1	&	0/\;\;8/2	&	0/10/0	    &	3/\;\;4/3	\\
                30  & \;\;600	&   0/10/0	    &	0/10/0	    &	2/6/2 	&	0/\;\;8/2 	&	2/7/1	&	2/3/5	&	1/\;\;8/1	&	0/10/0	    &	0/10/0	    &	2/\;\;8/0	\\
                    & \;\;800  	&   0/10/0	    &	0/10/0	    &	2/7/1 	&	0/\;\;9/1 	&	2/7/1	&	2/3/5	&	0/\;\;9/1	&	0/\;\;9/1	&	0/10/0	    &	2/\;\;5/3	\\
                    & 1,000  	&	0/10/0	    &	0/10/0	    &	1/7/2 	&	0/\;\;9/1 	&  	2/6/2	&	2/3/5	&	0/10/0	    &	0/\;\;9/1	&	0/10/0	    &	3/\;\;5/2	\\ \hline
                    & \;\;200  	&	0/\;\;9/1	&	0/\;\;8/2	&	2/6/2 	&	5/\;\;2/3 	&	0/6/4	&	1/3/6	&	0/\;\;8/2	&	0/\;\;7/3	&	0/\;\;8/2	&	0/\;\;8/2	\\
                    & \;\;400  	&	0/10/0	    &	0/10/0	    &	4/5/1 	&	4/\;\;4/2 	&	0/8/2	&	6/4/0	&	1/\;\;7/2	&	0/\;\;9/1	&	0/10/0	    &	2/\;\;5/3	\\
                50  & \;\;600	&	0/10/0	    &	0/10/0	    &	4/6/0 	&	2/\;\;7/1 	&	1/7/2	&	4/4/2	&	1/\;\;9/0	&	0/10/0	    &	0/10/0	    &	2/\;\;8/0	\\
                    & \;\;800  	&	0/10/0	    &	0/10/0	    &	3/7/0 	&	0/\;\;9/1 	&	2/8/0	&	3/5/2	&	0/10/0	    &	0/10/0	    &	0/10/0	    &	0/10/0	    \\
                    & 1,000  	&	0/10/0	    &	0/10/0	    &	2/8/0 	&	0/10/0    	&	2/8/0	&	2/5/3	&	0/10/0	    &	0/10/0	    &	0/10/0	    &	0/10/0	    \\ \hline
                    & \;\;200  	&	0/\;\;8/2	&	0/\;\;8/2	&	3/4/3 	&	8/\;\;1/1 	&	0/7/3	&	1/5/4	&	0/\;\;8/2	&	0/\;\;8/2	&	0/\;\;8/2	&	0/\;\;8/2	\\
                    & \;\;400  	&	2/\;\;7/1	&	0/10/0	    &	6/4/0 	&	5/\;\;3/2 	&	0/7/3	&	4/4/2	&	0/\;\;9/1	&	0/\;\;9/1	&	0/\;\;9/1	&	0/\;\;9/1	\\
                100 & \;\;600	&	2/\;\;7/1	&	0/10/0	    &	6/4/0 	&	2/\;\;5/3 	&	1/7/2	&	4/6/0	&	1/\;\;9/0	&	0/\;\;9/1	&	0/10/0	    &	1/\;\;8/1	\\
                    & \;\;800  	&	1/\;\;8/1	&	0/10/0	    &	5/5/0 	&	0/\;\;8/2 	&	1/7/2	&	4/6/0	&	1/\;\;8/1	&	0/10/0	    &	0/10/0	    &	2/\;\;8/0	\\
                    & 1,000  	&	2/\;\;8/0	&	0/10/0	    &	4/5/1 	&	0/\;\;8/2 	&   1/6/3	&	1/7/2	&	1/\;\;9/0	&	0/10/0	    &	0/10/0	    &	1/\;\;8/1	\\ \hline
            \end{tabular}
        }
    \end{minipage}
    }
\end{table}

\begin{table}[t]
    \centering
    \caption{Computational time [sec] at $1,000$ function evaluations averaged for $31$ trials and all problems for $D\in\{30,50,100\}$.}
    \vspace{-2mm}
    {\tabcolsep = 1mm
    \label{tab:runtime}
    \begin{minipage}[t]{0.495\textwidth}
        \centering
        \scalebox{0.925}{
            \begin{tabular}{c|ccccccccccc}\hline\hline
                $D$ & \hspace{-2mm}\begin{tabular}{c} GL- \\[-1mm] SADE \end{tabular}\hspace{-2mm} & IDRCEA & \hspace{-2mm}\begin{tabular}{c} TS- \\[-1mm] DDEO \end{tabular}\hspace{-2mm} & \hspace{-2mm}\begin{tabular}{c} MHS- \\[-1mm] QPSO \end{tabular}\hspace{-2mm} & \hspace{-2mm}\begin{tabular}{c} SADE- \\[-1mm] ATDSC \end{tabular}\hspace{-2mm} & \hspace{-2mm}\begin{tabular}{c} Auto \\[-1mm] SAEA \end{tabular}\hspace{-2mm} \\\hline
                30  &	1.832E+02	& 5.593E+01	& 1.445E+02	& 2.218E+02	& 7.795E+01	& 1.110E+02	 \\
                50  &	6.552E+02	& 1.041E+02	& 1.935E+02	& 1.717E+02	& 8.352E+01	& 1.186E+02	 \\
                100 &	1.089E+03	& 2.506E+02	& 2.869E+02	& 2.772E+02	& 1.056E+02	& 1.797E+02	 \\\hline\hline
                $D$ & \hspace{-2mm}\begin{tabular}{c} CAL- \\[-1mm] SAPSO \end{tabular}\hspace{-2mm} & EAPSO & HESNFO & \hspace{-2mm}\begin{tabular}{c} DSP- \\[-1mm] SAEA \end{tabular}\hspace{-2mm} & \ours \\\hline
                30  &	2.308E+04	& 6.008E+02	& 6.679E+01	& 4.227E+03	& 4.737E+03	\\
                50  &	1.485E+04	& 2.221E+02	& 8.508E+01	& 9.017E+04	& 6.124E+03	\\
                100 &	3.591E+04	& 5.610E+02	& 1.179E+02	& 2.110E+04	& 9.573E+03	\\\hline
            \end{tabular}
        }
    \end{minipage}
    }
\end{table}

\section{Discussion} \label{sec:dis}

This section evaluates the effectiveness of the main components of \ours and to understand its behavior and general versatility. First, we discuss the effect of multi-objectivization and the use of an MOEA in the surrogate model adaptation. Subsequently, we investigate the effectiveness of using the uncertainty in the ensemble. Then, we examine whether our approach is also effective within the Bayesian Optimization (BO) framework. We also compare our approach with the existing surrogate modeling methods. Lastly, we apply \ours to real-world problems. Additionally, we report four discussions in the Supplementary material: comparison of the search strategies (Section S-VII), sensitivity analysis of hyperparameters of model construction and evolutionary optimization (Sections S-VIII\&S-IX, respectively), and investigation of the effect of the ensemble size in the Pareto set on the performance (Section S-X). Experimental settings follow the previous section.

\subsection{Impact of Multi-objectivization in the Adaptation Phase} \label{sec:dis-MOP}

\ours optimizes an MOP of minimizing the approximation error and model complexity to adapt the RBFN structures. This subsection evaluates the effectiveness of multi-objectivization by comparing the performances of \ours and its SOP version variants. Two variants named ``Acc-only'' and ``Smp-only'' minimize the approximation error and model complexity, respectively, using GA instead of NSGA-II. In other words, they optimize only the approximation accuracy or simplicity of surrogate models, respectively. Specifically, the pseudo-code of the variant is given in Algorithm S-I of the Supplementary material, where the changes from Algorithm~\ref{alg:our} are underlined. In Lines 9 and 13, the solutions in the population $\mathcal{P}_{M}$ are sorted based on the focused single objective. During prescreening of the DE offspring process (Line 17), the variant uses the best surrogate model in $\mathcal{P}_{M}$ instead of the ensemble surrogate model. Successively, an offspring with the best pseudo-fitness value calculated with this surrogate model is selected for FE.

TABLE~\ref{tab:sigs_dis} a) presents the statistical results of the Wilcoxon signed-rank test summarized as the counts of ``$+\hspace{-1pt}/\hspace{-1pt}-\hspace{-1pt}/\hspace{-1pt}\sim$'' at $200$, $400$, $600$, $800$, and $1,000$ FEs. Each sign ``$+$,'' ``$-$,'' and ``$\sim$'' indicates that the performance of a variant is statistically better than, worse than, and competitive with that of \our, respectively. On comparing the ``Acc-only'' variant and \our, the variant outperformed \ours under $600$ FEs but underperformed \ours with $800$ and $1,000$ FEs with $D \in \{30, 50\}$, judging from the numbers of ``$+$'' and ``$-$.'' This demonstrates the need to consider minimizing model complexity in \our, otherwise an SAEA with adaptation only considering approximation error minimization causes the premature convergence of the performance. When $D$ increased to $100$, the difference between the number of ``$+$'' and ``$-$'' became smaller, although the number of ``$-$'' exceeded that of ``$+$''. Therefore, although \ours should deal with both objectives, the approximation error improves the performance of \ours well in high-dimensional cases. Conversely, the other variant ``Smp-only'' significantly underperformed \ours in all cases shown in TABLE~\ref{tab:sigs_dis} a). This indicates that the surrogate model adaptation considering only the minimization of model complexity cannot contribute to the performance improvement of an SAEA.

\begin{table}[t]
    \centering
    \caption{Significant differences regarding findings for ``$+\hspace{-1pt}/\hspace{-1pt}-\hspace{-1pt}/\hspace{-1pt}\sim$'' in comparison between \ours and its variants. Results were obtained in Sections \textnormal{a}) \ref{sec:dis-MOP}, \textnormal{b}) \ref{sec:dis-MOEA}, and \textnormal{c}) \ref{sec:dis-IC}.}
    \vspace{-2mm}
    {\tabcolsep = 1.4mm
    \label{tab:sigs_dis}
    \begin{minipage}[t]{0.495\textwidth}
        \centering
        \scalebox{0.9}{
            \begin{tabular}{c|c|cc|cc|cc}\hline\hline
                \multicolumn{2}{c|}{} & \multicolumn{2}{c|}{a) Section~\ref{sec:dis-MOP}} & \multicolumn{2}{c|}{b) Section~\ref{sec:dis-MOEA}} & \multicolumn{2}{c}{c) Section~\ref{sec:dis-IC}} \\\hline
                $D$ & FEs       & \hspace{-2mm}\begin{tabular}{c} vs \\[-1mm] Acc-only \end{tabular}\hspace{-2mm} & \hspace{-2mm}\begin{tabular}{c} vs \\[-1mm] Smp-only \end{tabular}\hspace{-2mm} & \hspace{-2mm}\begin{tabular}{c} vs \\[-1mm] NoMOEA \end{tabular}\hspace{-2mm} & \hspace{-2mm}\begin{tabular}{c} vs \\[-1mm] NoSigma \end{tabular}\hspace{-2mm} & \hspace{-2mm}\begin{tabular}{c} vs \\[-1mm] Mean \end{tabular}\hspace{-2mm} &  \hspace{-2mm}\begin{tabular}{c} vs \\[-1mm] EI \end{tabular}\hspace{-2mm} \\\hline
                    & \;\;200  	&   6/0/4	    &	0/3/7	    &   0/1/9	    &	0/2/8	    &   2/1/\;\;7	&	1/0/\;\;9	\\
                    & \;\;400  	&   5/0/5	    &	0/6/4	    &   0/3/7	    &	0/5/5	    &   3/0/\;\;7	&	0/2/\;\;8	\\
                30  & \;\;600	&   3/2/5	    &	0/8/2	    &   0/5/5	    &	1/7/2	    &   2/3/\;\;5	&	0/5/\;\;5	\\
                    & \;\;800  	&   0/4/6	    &	0/7/3	    &   1/4/5	    &	2/7/1	    &   0/3/\;\;7	&	1/2/\;\;7	\\
                    & 1,000  	&	0/5/5	    &	0/7/3       &	1/3/6	    &	1/7/2       &	0/3/\;\;7	&	0/0/10      \\ \hline
                    & \;\;200  	&	5/0/5	    &	0/3/7	    &	0/2/8	    &	0/4/6	    &	0/0/10	    &	1/1/\;\;8	\\
                    & \;\;400  	&	6/1/3	    &	0/6/4	    &	0/2/8	    &	0/4/6	    &	1/0/\;\;9	&	0/1/\;\;9	\\
                50  & \;\;600	&	5/1/4	    &	0/8/2	    &	0/4/6	    &	0/7/3	    &	2/3/\;\;5	&	1/6/\;\;3	\\
                    & \;\;800  	&	1/3/6	    &	0/8/2	    &	1/6/3	    &	0/7/3	    &	0/3/\;\;7	&	0/5/\;\;5	\\
                    & 1,000  	&	1/4/5	    &	0/7/3	    &	1/4/5	    &	0/7/3	    &	1/3/\;\;7	&	0/2/\;\;8   \\ \hline
                    & \;\;200  	&	0/4/6	    &	0/4/6	    &	1/1/8	    &	2/3/5	    &	2/1/\;\;7	&	1/0/\;\;9	\\
                    & \;\;400  	&	4/4/2	    &	0/3/7	    &	0/3/7	    &	0/4/6	    &	0/0/10	    &	0/1/\;\;9	\\
                100 & \;\;600	&	4/5/1	    &	0/7/3	    &	0/4/6	    &	0/5/5	    &	2/0/\;\;8	&	0/2/\;\;8	\\
                    & \;\;800  	&	3/5/2	    &	1/8/1	    &	0/3/7	    &	0/4/6	    &	3/3/\;\;4	&	0/3/\;\;7	\\
                    & 1,000  	&	4/5/1	    &	0/8/2       &	0/4/6	    &	0/6/4       &	2/4/\;\;4	&	0/6/\;\;4   \\ \hline
            \end{tabular}
        }
    \end{minipage}
    }
\end{table}

\subsection{Impact of MOEA in the Adaptation Phase} \label{sec:dis-MOEA}

\ours uses an MOEA, i.e., NSGA-II, to optimize the aforementioned MOP in the surrogate model adaptation phase. Now, we verify the necessity of an MOEA in optimizing this MOP and the adequacy of its design. We compare \ours to two variants here. One variant ``NoMOEA'' randomly produces the same number of surrogate models using LHS as the total number of surrogate models generated in the original \our, i.e., ${N}_{M} + {N}_{M} \times {\omega}_{\max} = 110$. Then, the Pareto set $\mathcal{PS}$ for the surrogate model ensemble is created by applying NDCD to the generated $110$ surrogate model. The other procedures inherit from the original \our. Comparison with this variant can clarify the necessity of an MOEA in the surrogate model adaptation phase. The other variant ``NoSigma'' is prepared to show the need for the second decision variable ${x}_{M,2}=\sigma$ (the spread parameter of an RBFN) for the surrogate model adaptation. Note that the first decision variable ${x}_{M,1}=n$ (number of hidden layer nodes) clearly contributes to the minimization of the model complexity. Therefore, in this variant, ${x}_{M,2}=\sigma$ is fixed to $\max(\mathit{dist}(\mathcal{D}_{\mathit{train}})) \times {(|\mathcal{D}_{\mathit{train}}| D)}^{-1/D}$, which is a recommended and representative setting in RBFN \cite{Ikeguchi2025-pn}.

We summarized the Wilcoxon signed-rank test results in TABLE~\ref{tab:sigs_dis} b), where ``$+$,'' ``$-$,'' and ``$\sim$'' mean the superiority, the inferiority, and the competitiveness of the variant to \our, respectively. First, we can detect that the number of ``$-$'' was larger than that of ``$+$'' in the ``vs NoMOEA'' column and the maximum number of ``$-$'' was six ($D=50$, $800$ FEs). These results proved that an MOEA contributed to the performance improvement of \our. The evolutionary pressure of NSGA-II helped \ours generate dominant but diverse surrogate models in the objective space. These surrogate models provided a good ensemble model. Compared to ``NoSigma,'' \ours outperformed ``NoSigma'' as a whole, which can be confirmed by the much larger number of ``$-$'' than that of ``$+$.'' The adaptive control of ${x}_{M,2}=\sigma$ is indispensable to draw suitable approximated fitness landscapes as the centers of RBFNs move with the change of search dynamics.

\subsection{Impact of Uncertainty in the Ensemble Surrogate Model} \label{sec:dis-IC}

In the prescreening phase of the DE offspring, \ours employs LCB as the infill criterion. LCB utilizes the uncertainty, i.e., the unbiased estimation of the standard deviation obtained by the ensemble surrogate model shown in Eq.~(\ref{eq:sd}). We first investigate the effectiveness of using uncertainty in this subsection. Specifically, the ``Mean'' variant is prepared and uses only the mean of outputs of surrogate models in $\mathcal{PS}$ during the offspring prescreening described in Eq.~(\ref{eq:mean}). We arranged the other variant, named ``EI,'' to review the effect of the infill criterion on the performance of \our. It uses the EI infill criterion for prescreening, calculated as follows:
\begin{equation}
	\mathit{EI} = \left({f}_{\min} - \bar{f}(\bm{x})\right)\Phi(Z) + \hat{s}\left(\hat{f}(\bm{x})\right)\varPhi (Z), \label{eq:EI}
\end{equation}
where $\bar{f}(\bm{x})$ and $\hat{s}\left(\hat{f}(\bm{x})\right)$ follow Eqs.~(\ref{eq:mean}) and (\ref{eq:sd}), respectively, ${f}_{\min}$ is the best fitness value found so far, $Z=\frac{{f}_{\min} - \bar{f}(\bm{x})}{\hat{s}\left(\hat{f}(\bm{x})\right)}$, and $\Phi(\cdot)$ and $\varPhi(\cdot)$ are the cumulative distribution function and probability density function of standard normal distribution, respectively. The DE offspring with the largest EI value is selected for the FE in this variant.

TABLE~\ref{tab:sigs_dis} c) provides the summary of Wilcoxon signed-rank test results between \ours and two variants. Every sign ``$+$,'' ``$-$,'' and ``$\sim$'' indicates that the performance of a variant is statistically better than, statistically worse than, and competitive with that of \our, respectively. The performance of the ``Mean'' variant was competitive to or better than \ours before $600$ FEs ($D\in\{30,50\}$) and $800$ FEs, which implies that using only the mean of outputs of surrogate models is effective to improve convergence speed. In addition, the effectiveness of the ensemble of surrogate models itself is confirmed because the variant is competitive with \our, which was superior to the state-of-the-art SAEAs. However, as the superior performance of the original \ours after $800$ FEs ($D\in\{30,50\}$) and $1,000$ FEs can be seen in the table, the utilization of uncertainty contributed to sustainable performance improvement of \ours while escaping premature convergence. This is because LCB can find unexplored regions. Moreover, leveraging the uncertainty enhances the robustness of surrogate model prediction. In the comparison of ``EI'' and \our, while they were competitive at $D=30$, \ours (LCB) outperformed ``EI'' with $D\in\{50,100\}$. While both LCB and EI contributed to finding better solutions in lower-dimensional cases, the infill criterion with relatively strong exploitation ability, i.e., LCB, is preferred in higher-dimensional cases where the number of FEs per decision variable is limited.

\subsection{Impact of the Proposed Ensemble Approach and Emulated LCB on the Bayesian Optimization Framework} \label{sec:dis-BO}

We evaluate whether the proposed ensemble approach and LCB calculated with model disagreement are also effective for BO with Kriging. BO is another strong approach to EOPs alongside SAEAs. We compare the performance of the basic BO and its variant with our ensemble approach. The DACE model \cite{Lophaven2002-bf} with the linear regression model and the Gaussian correlation model was employed for Kriging. For the acquisition function, LCB calculated from the process variance was used in the basic BO, unlike that used in \our. Conversely, in the prepared variant, the prediction error and the degree of smoothness of the approximated fitness landscape derived by DACE models are minimized by NSGA-II. The decision variables of NSGA-II are the correlation parameters ${\theta}_{d}$ ($d=1,2,\dots,D$) of the Gaussian correlation function $\exp\left(-{\theta}^{2}_{d}\left( {x}_{i,d} - {x}_{j,d} \right)^2\right)$ (for the $d$-th dimension of two points $\bm{x}_{i}$ and $\bm{x}_{j}$). The correlation parameters determine the strength of correlation between data points, affecting the degree of smoothness of the approximated fitness landscape. The prediction error is calculated in the same manner as \our, and the degree of smoothness is expressed as the mean of ${\theta}_{d}$ for $d=1,2,\dots,D$. After Pareto-optimal DACE models are obtained, the variant constructs their ensemble model like \our. The emulated LCB used in the variant is calculated using the unbiased estimation of the standard deviation.

TABLE~\ref{tab:sigs_BO} shows the summary of results at $1,000$ FEs for $D \in \{30,50,100\}$, where the results of the Wilcoxon signed-rank test are reported with ``$+$,'' ``$-$,'' or ``$\sim$,'' which stands for the original BO significantly outperformed, significantly underperformed, or was competitive to the variant, respectively. The variant with the proposed ensemble model statistically outperformed the original BO on $17$ problem instances and obtained a much better average rank. This demonstrates that the proposed ensemble approach and emulated LCB are also effective on the BO framework and Kriging. Moreover, the superiority of the proposed ensemble approach over existing surrogate modeling methods was confirmed.

\begin{table}[t]
    \centering
    \caption{The summary of results at $1,000$ fitness evaluations obtained in the comparison between Bayesian Optimization (BO) using the lower confidence bound (LCB) and its variant that utilizes ensemble models and emulates LCB as \ours does.}
    \vspace{-2mm}
    {\tabcolsep = 1.5mm
    \label{tab:sigs_BO}
    \begin{minipage}[t]{0.495\textwidth}
        \centering
        \scalebox{1}{
            \begin{tabular}{c|cc}\hline\hline
                 & BO using LCB & BO using the ensemble model \\\hline
                $+/-/\sim$	&		0/17/13		    &		- \\
                Ave. Rank	&	\worst	1.900 		&	\win	1.100  \\ \hline
            \end{tabular}
        }
    \end{minipage}
    }
\end{table}

\subsection{Comparison with Existing Surrogate Modeling Methods} \label{sec:dis-SMM}

Next, we verified the contribution of our proposed surrogate modeling method over existing ones to the optimization performance. We compared the performance of nine surrogate modeling methods including ours on the \ours framework, i.e., we changed Lines~\ref{alg-l:adaptation-begin}-\ref{alg-l:MOEA-end} of Algorithm~\ref{alg:our} to the following eight surrogate modeling methods for the variants; RBFN \cite{Park1991-et}, Kriging \cite{Lophaven2002-bf}, PR \cite{Myers1995-cu}, and k Nearist Neighbor (kNN, L1-exploitation style \cite{Xie2024-qk}, a classification model) as the basic surrogate models used in the compared algorithms in Section~\ref{sec:exp}, ELDR \cite{Harada2023-wp} and XGBoost \cite{Chen2016-hf} as the recently proposed surrogate models, and an ensemble surrogate model of RBFN with different RBFs (Ensemble-RBFs) and that of RBFN, Kriging, and PR (Ensemble-Models), which are the bagging-style ensemble models employed in HESNFO and DSP-SAEA, respectively. XGBoost is also an ensemble model (boosting-style). The experimental design is identical to the main experiment.

TABLE~\ref{tab:sigs_SMM} summarizes the Wilcoxon signed-rank test results as well as the average rank for $10$ problems with the setting of $D\in\{30,50,100\}$, wherein ``$+$,'' ``$-$,'' and ``$\sim$'' correspond to the significantly high, significantly low, and competitive performance of the compared algorithm to \our. In the table, \ours statistically outperformed the four basic surrogate models on most problem instances. In the comparison with RBFN, the effectiveness of the automatic adaptation of RBFN performed by \ours can be confirmed. Since the average rank was the worst despite the presence of two ``$+$''s, it lacked the robustness of performance. Moreover, the performance of \ours was significantly superior to that of the \ours variants with ELDR or XGBoost although the number of ``$-$'' slightly decreased, which implies the importance of considering the model complexity during fitting of surrogate models. Furthermore, \ours statistically outperformed Ensemble-RBFs and Ensemble-Models, thanks to its elaborate adaptation of ensemble elements and the effectiveness of our infill criterion. These results demonstrated the superiority of the proposed surrogate modeling method over existing ones.

\begin{table}[t]
    \centering
    \caption{The summary of results at $1,000$ fitness evaluations obtained in the comparison with existing surrogate modeling methods.}
    \vspace{-2mm}
    \label{tab:sigs_SMM}
    \centering
    \begin{minipage}[t]{0.495\textwidth}
        \centering
            \scalebox{1}{
                \begin{tabular}{c|ccccccccccc}\hline\hline
                    &		RBFN		&		Kriging		&		PR		&		kNN		&		ELDR		\\ \hline
                    $+/-/\sim$	                &		0/24/6		&		2/28/0		&		0/30/0		&		0/29/1		&		0/26/4		\\
                    Ave. Rank	                &		5.367 		&	\worst	7.867 		&		5.867 		&		6.400 		&		4.800 		\\ \hline\hline
                    &		XGBoost		&		\hspace{-2mm}\begin{tabular}{c} Ensemble- \\[-1mm] RBFs \end{tabular}\hspace{-2mm}		&		\hspace{-2mm}\begin{tabular}{c} Ensemble- \\[-1mm] Models \end{tabular}\hspace{-2mm}		&		\ours	\\ \hline
                    $+/-/\sim$	                &		0/23/7		&		0/28/2		&		2/24/4		&		-	\\
                    Ave. Rank	                &		4.067 		&		5.700 		&		3.633 		&	\win	1.300 	\\ \hline
                \end{tabular}
            }
    \end{minipage}
\end{table}

\subsection{Real-world Application} \label{sec:dis-RW}

Finally, we apply \ours to real-world problems. We used the function suite proposed in the IEEE CEC 2011 Competition on Testing Evolutionary Algorithms on Real-World Optimization Problems \cite{Das2010-xj}. This suite comprises $22$ problems from a variety of scientific fields including signal processing (F1, F7, and F10), materials science (F2, F5, and F6), chemical engineering (F3 and F4), energy systems (F8, F9, and F11), and aerospace engineering (F12 and F13). The problem dimension varies from $1$ to $216$. The same compared algorithms were employed as the main experiment in Section~\ref{sec:exp}. The maximum number of FEs was set to $1,000$ and the performance is reported with the average fitness values of $31$ independent trials. The Wilcoxon signed-rank test was again used for each pair of \ours and a compared algorithm. These test results are reported with ``$+$,'' ``$-$,'' or ``$\sim$,'' which means the compared algorithm significantly outperformed \our, significantly underperformed \our, or we cannot find a significant difference, respectively.

We provide the fitness values at $1,000$ FEs in TABLE S-VIII of the Supplementary material with standard deviations in parentheses, where the best and worst results are marked in green bold and pink italics, respectively. The summary of results is given in TABLE~\ref{tab:data_CEC2011RW}. GL-SADE failed in running in the F6 problem. \ours derived no ``$+$,'' indicating the inferiority of \our, against CAL-SAPSO, EAPSO, HESNFO, and DSP-SAEA while \ours got ``$-$,'' indicating the superiority of \our, in all the problems in the comparisons with CAL-SAPSO or HESNFO. In the comparisons with SADE-ATDSC or AutoSAEA, results were both ``$+\hspace{-1pt}/\hspace{-1pt}-\hspace{-1pt}/\hspace{-1pt}\sim=4/8/10$.'' \ours was highly competitive to these two algorithms. However, AutoSAEA lacked stability in performance. For example, it derived both the best and worst results in TABLE S-VIII. DSP-SAEA, which was competitive to \ours for $D=30$ with the benchmark suite (TABLE~\ref{tab:sigs}), was the worst in six problem instances in TABLE S-VIII. Similarly, normal SAEAs for high-dimensional EOPs derived some worst values, indicating the high dependence of their performance on $D$. Conversely, \ours was the worst with no problems, demonstrating that the performance of \ours was stable thanks to the robust design of ensemble surrogate models. Furthermore, \ours was the best within a wide range of $D$, i.e., from $13$ to $140$, and derived the best average rank as shown in the last row of the table. We also provide the convergence curves of the performance in Fig. S-2 of the Supplementary material, which also demonstrates the good convergence speed and continuous performance improvement of \our. In summary, \ours outperformed the compared algorithms in many problems with different difficulties from wide engineering fields, confirming the effectiveness of \ours on various real-world applications.

\begin{table}[t]
    \renewcommand{\arraystretch}{1.1}
    \centering
    \caption{The summary of results at $1,000$ fitness evaluations on the CEC2011 Real-world Problem Suite.}
    \vspace{-2mm}
    {\tabcolsep = 1.5mm
    \label{tab:data_CEC2011RW}
    \begin{minipage}[t]{0.495\textwidth}
        \centering
        \scalebox{1}{
            \begin{tabular}{c|ccccccccccc}\hline\hline
                 & \hspace{-2mm}\begin{tabular}{c} GL- \\[-1mm] SADE \end{tabular}\hspace{-2mm} & IDRCEA & \hspace{-2mm}\begin{tabular}{c} TS- \\[-1mm] DDEO \end{tabular}\hspace{-2mm} & \hspace{-2mm}\begin{tabular}{c} MHS- \\[-1mm] QPSO \end{tabular}\hspace{-2mm} & \hspace{-2mm}\begin{tabular}{c} SADE- \\[-1mm] ATDSC \end{tabular}\hspace{-2mm} & \hspace{-2mm}\begin{tabular}{c} Auto \\[-1mm] SAEA \end{tabular}\hspace{-2mm} \\\hline
                $+/-/\sim$		                &	6/12/4	&	2/18/2	&	4/14/4	&	2/17/3	&	4/8/10	&	4/8/10	\\ 
                Ave. Rank		                &	5.500 	&	7.727 	&	5.591 	&	6.682 	&	3.273 	&	3.909 	\\ \hline\hline
                 & \hspace{-2mm}\begin{tabular}{c} CAL- \\[-1mm] SAPSO \end{tabular}\hspace{-2mm} & EAPSO & HESNFO & \hspace{-2mm}\begin{tabular}{c} DSP- \\[-1mm] SAEA \end{tabular}\hspace{-2mm} & \ours \\\hline
                $+/-/\sim$		                &	0/22/0	&	0/18/4	&		0/22/0		&	0/18/4	&		-           \\ 
                Ave. Rank		                &	8.182 	&	4.727 	&	\worst	9.455 	&	8.000 	&	\win	2.955   \\ \hline
            \end{tabular}
        }
    \end{minipage}
    }
\end{table}

\section{Conclusion} \label{sec:con}

This study proposed an SAEA with adaptation and ensemble mechanisms of the RBFN surrogate models, namely \our. In the adaptation phase, NSGA-II simultaneously minimizes the approximation error and model complexity to optimize the RBFN model structure. This enables \ours to automatically and flexibly design surrogate models with plausible but diverse degrees of smoothness of the approximated fitness landscapes while maintaining the approximation accuracy. Subsequently, the obtained surrogate models in the Pareto-optimal set are utilized to construct an ensemble model. \ours uses the LCB infill criterion to prescreen the offspring generated by DE. This mechanism improved both the approximation accuracy and robustness of the surrogate model.
As a result, \ours significantly outperformed existing state-of-the-art SAEAs on both benchmark and real-world problems. Intensive experiments between \ours and its variants showed the necessity of multi-objectivization and an MOEA in the automatic design of surrogate models. Consequently, this work presented that the evolutionary automatic design of Pareto-optimal surrogate models is a promising methodology for constructing a high-performing and robust ensemble model.

Future work includes designing a surrogate model complexity criterion that is effective for different types of ML models. The effectiveness of other types of ML models for surrogate models and MOEAs to automatically design surrogate models in the \ours framework should be evaluated. We also plan to reduce the computational time required for surrogate model adaptation because SAEAs with short computational time are welcomed in financially expensive optimization problems. Moreover, we are also motivated to improve the ensemble method. Using the model disagreement for LCB may not work when there are many extremely inaccurate models or when all models make the same prediction. We will develop a mechanism to select only valid surrogate models rather than using all of them. Additionally, it is necessary to theoretically clarify the convergence behavior of \ours as well as error bounds and uncertainty convergence of the proposed ensemble surrogate model. This is partially because our emulated uncertainty using model disagreement is not formally grounded in probability theory, unlike Kriging. Furthermore, \ours should be extended for expensive MOPs and expensive constrained SOPs/MOPs \cite{Nishihara2024-tu}, as well as extremely expensive cases wherein only almost $200$ FEs are allowed.


%





\ifCLASSOPTIONcaptionsoff
  \newpage
\fi



%

\bibliographystyle{IEEEtran}
\bibliography{myref}


%

\begin{IEEEbiography}[{\includegraphics[width=1in,height=1.25in,clip,keepaspectratio]{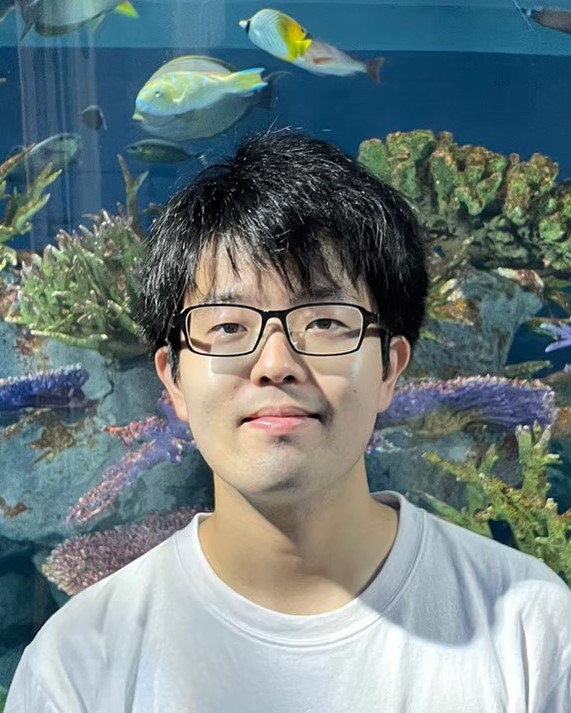}}]{Kei Nishihara}
    (Member, IEEE) received the B.Eng., M.Eng., and PhD degrees from Yokohama National University, Yokohama, Japan, in 2020, 2022, and 2025, respectively. He served as an assistant professor at the Graduate School of Engineering, Muroran Institute of Technology, Muroran, Japan, and is currently an assistant professor at Yokohama National University, Yokohama, Japan. His current research interest includes adaptation in evolutionary algorithms and surrogate-assisted evolutionary algorithms.
\end{IEEEbiography}

\begin{IEEEbiography}[{\includegraphics[width=1in,height=1.25in,clip,keepaspectratio]{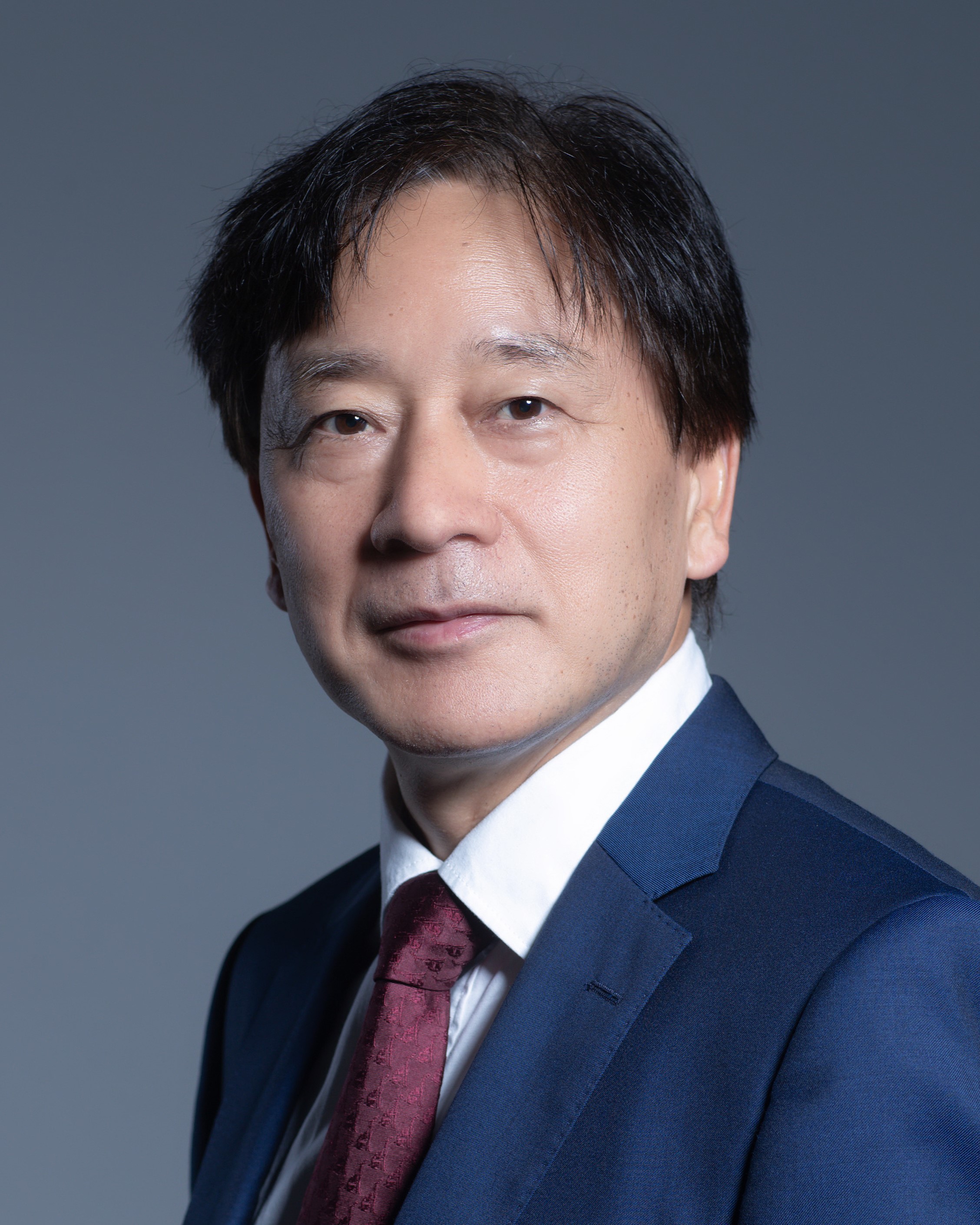}}]{Yaochu Jin}
    (Fellow, IEEE) received the BSc, MSc, and PhD degrees in automatic control from Zhejiang University, Hangzhou, China, in 1988, 1991, and 1996, respectively, and the Dr.-Ing. degree from Ruhr-University Bochum, Bochum, Germany, in 2001. 
    
    He is currently a Chair Professor of AI with the School of Engineering, Westlake University, Hangzhou, China, leading the Trustworthy and General AI Laboratory. He was Alexander von Humboldt Professor for AI with the Faculty of Technology, Bielefeld University, Germany, and Surrey Distinguished Chair of Computational Intelligence, University of Surrey, U.K. He was a Finland Distinguished Professor and a Changjiang Distinguished Visiting Professor in China. His main research interests include data-driven evolutionary optimization, trustworthy machine learning, multiobjective evolutionary learning, and evolutionary developmental systems.

    Dr. Jin is the recipient of the 2025 IEEE Frank Rosenblatt Award, the 2018, 2021, and 2024 IEEE \textsc{Transactions on Evolutionary Computation} Outstanding Paper Award, and the 2015, 2017, and 2020 IEEE Computational Intelligence Magazine Outstanding Paper Award. He has been named ``Highly Cited Researcher'' since 2019 by Clarivate Analytics. He was the President of the IEEE Computational Intelligence Society in 2024-2025 and is currently the Editor-in-Chief of \textit{Complex \& Intelligent Systems}. He is a member of Academia Europaea.
\end{IEEEbiography}

\begin{IEEEbiography}[{\includegraphics[width=1in,height=1.25in,clip,keepaspectratio]{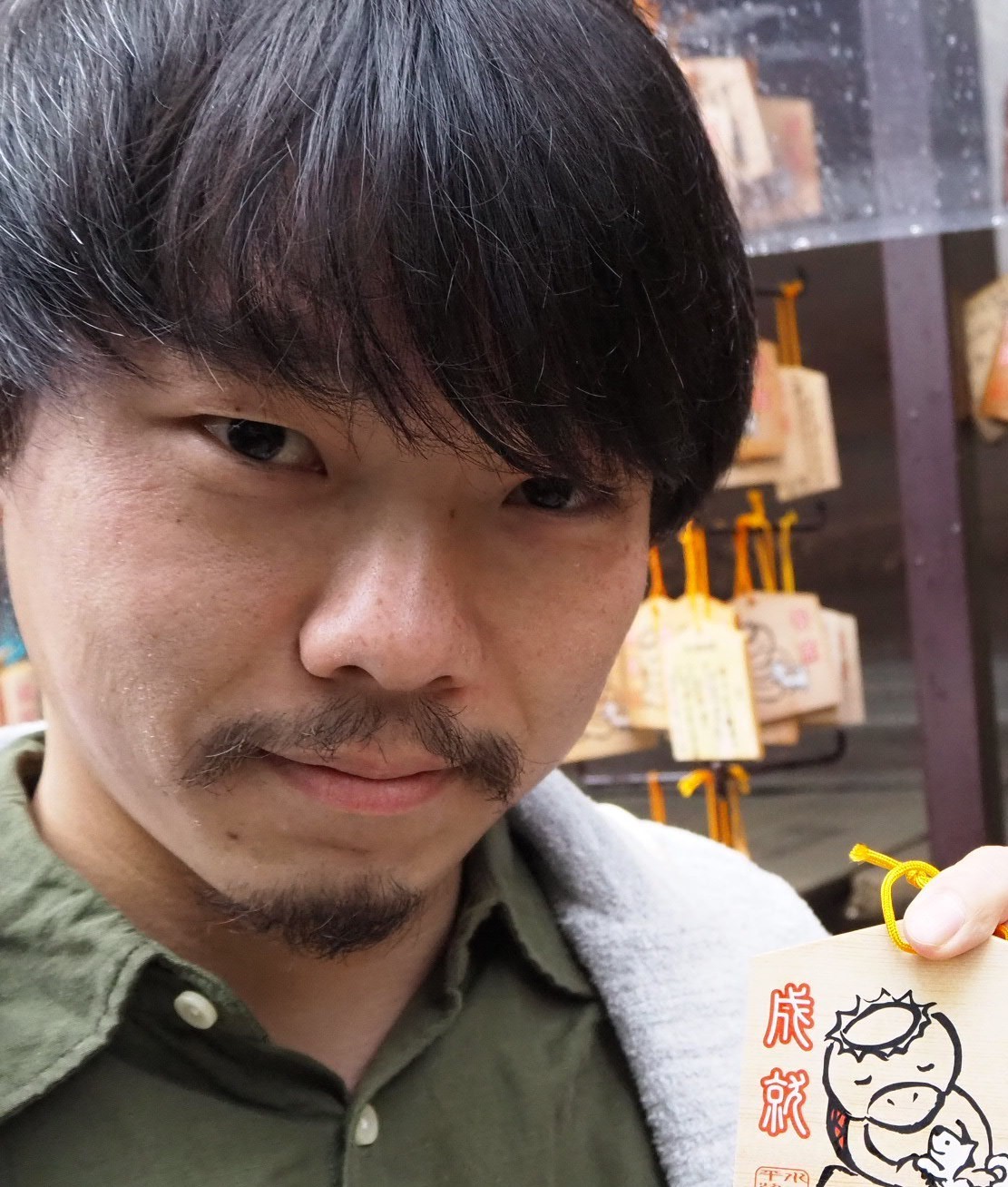}}]{Masaya Nakata}
    (Member, IEEE) received a Ph.D. degree in informatics from The University of Electro-Communications, Japan, in 2016. He is currently an Associate Professor at the Faculty of Engineering, Yokohama National University, Yokohama, Japan. He has been mainly working on evolutionary machine learning, and data mining, specifically, theoretical analysis of evolutionary rule-based learning. Since 2019, he has been focusing his research on surrogate-assisted evolutionary algorithms. His contributions have been published through more than 30 journal articles and 60 conference papers, such as the IEEE \textsc{Transactions on Evolutionary Computation}, GECCO, and PPSN.
\end{IEEEbiography}




\includepdf[pages=-]{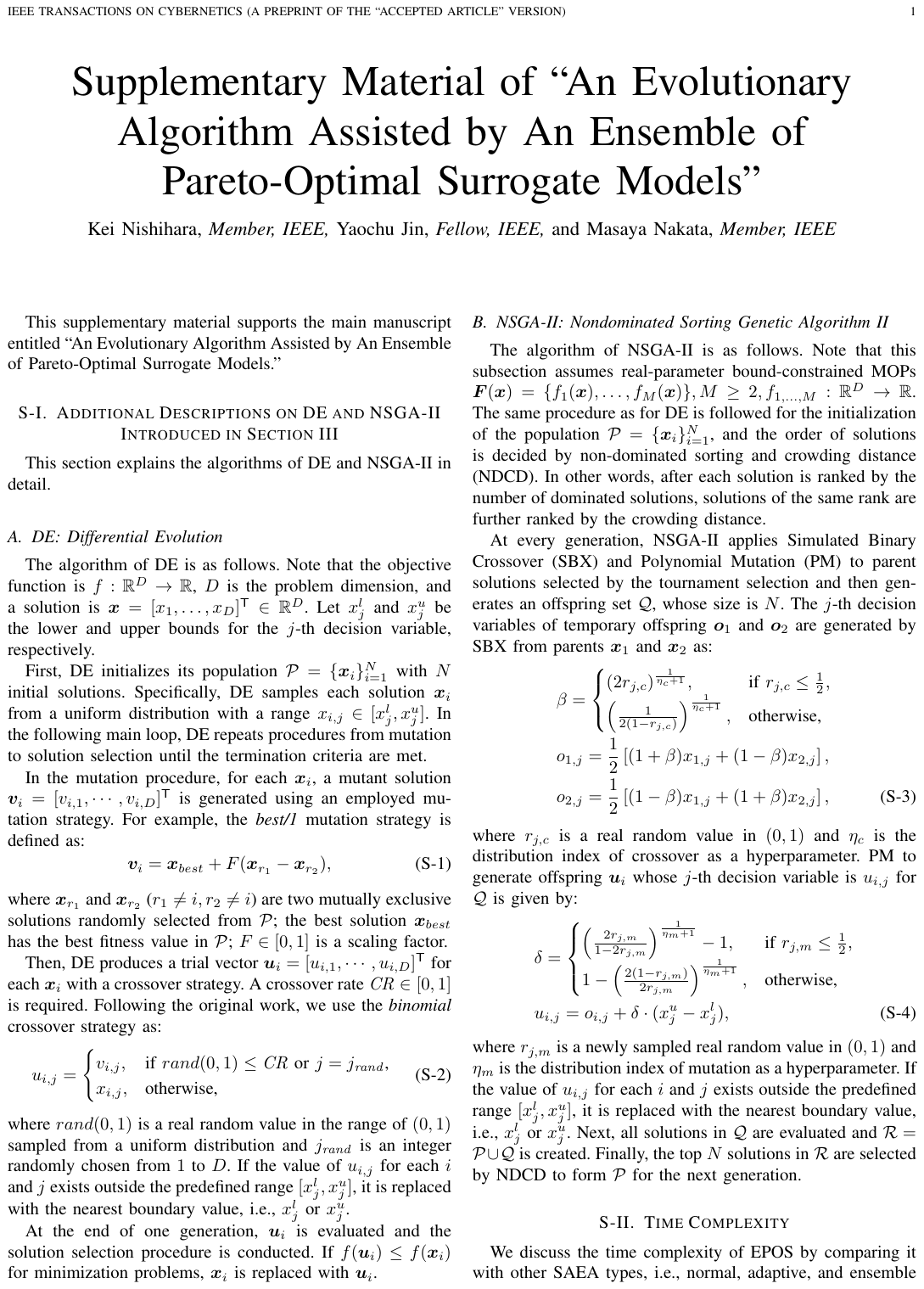}

\end{document}